\documentclass[pdflatex,sn-nature]{sn-jnl}

\usepackage{graphicx}%
\usepackage{multirow}%
\usepackage{amsmath,amssymb,amsfonts}%
\usepackage{amsthm}%
\usepackage{mathrsfs}%
\usepackage[title]{appendix}%
\usepackage{xcolor}%
\usepackage{textcomp}%
\usepackage{manyfoot}%
\usepackage{booktabs}%
\usepackage{algorithm}%
\usepackage{algorithmicx}%
\usepackage{algpseudocode}%
\usepackage{listings}%
\usepackage{tcolorbox}
\usepackage{makecell}
\usepackage{tabularx}
\usepackage{tabularray}
\usepackage{array}
\usepackage{hhline}
\usepackage{caption}
\usepackage{longtable}

\usepackage{xstring}

\newcounter{supmethod}
\renewcommand{\thesupmethod}{M\arabic{supmethod}}

\newcommand{\supmethod}[2]{%
  \refstepcounter{supmethod}%
  \subsection*{Supplementary Method \thesupmethod. #1}%
  \label{#2}%
}

\theoremstyle{thmstyleone}%
\theoremstyle{thmstyletwo}%

\theoremstyle{thmstylethree}%

\begin{document}

\title[Evaluating Clinical Competencies of Large Language Models with a General Practice Benchmark]{Evaluating Clinical Competencies of Large Language Models with a General Practice Benchmark}


\author[1]{\fnm{Zheqing} \sur{Li}} \equalcont{These authors contributed equally to this work.}
\author[2]{\fnm{Yiying} \sur{Yang}} \equalcont{These authors contributed equally to this work.}
\author[1]{\fnm{Jiping} \sur{Lang}} \equalcont{These authors contributed equally to this work.} 
\author*[2]{\fnm{Wenhao} \sur{Jiang}}\email{cswhjiang@gmail.com} \equalcont{These authors contributed equally to this work.}
\author*[1]{\fnm{Junrong} \sur{Chen}}\email{chenjr5@mail.sysu.edu.cn} \equalcont{These authors contributed equally to this work.}

\author[2]{\fnm{Yuhang} \sur{Zhao}}
\author[1]{\fnm{Shuang} \sur{Li}}
\author[2]{\fnm{Dingqian} \sur{Wang}}
\author[1]{\fnm{Zhu} \sur{Lin}}
\author[1]{\fnm{Xuanna} \sur{Li}}
\author[1]{\fnm{Yuze} \sur{Tang}}
\author[3]{\fnm{Jiexian} \sur{Qiu}}
\author[3]{\fnm{Xiaolin} \sur{Lu}}
\author[3]{\fnm{Hongji} \sur{Yu}}
\author[1]{\fnm{Shuang} \sur{Chen}}
\author[1]{\fnm{Yuhua} \sur{Bi}}
\author[1]{\fnm{Xiaofei} \sur{Zeng}}
\author[1]{\fnm{Yixian} \sur{Chen}}

\author*[4,5]{\fnm{Lin} \sur{Yao}}\email{yaolin@mail.sysu.edu.cn}

\affil[1]{\orgname{The Sixth Affiliated Hospital of Sun Yat-sen University}, \orgaddress{\city{Guangzhou}, \postcode{510655}, \state{Guangdong}, \country{China}}}

\affil[2]{\orgname{Guangdong Laboratory of Artificial Intelligence and Digital Economy (SZ)}, \orgaddress{\city{Shenzhen}, \postcode{518107}, \state{Guangdong}, \country{China}}}

\affil[3]{\orgname{Xinyi People's Hospital}, \orgaddress{\city{Xinyi}, \postcode{610101}, \state{Guangdong}, \country{China}}}

\affil[4]{\orgname{The Fifth Affiliated Hospital of Sun Yat-sen University}, \orgaddress{\city{Zhuhai}, \postcode{519000}, \state{Guangdong}, \country{China}}}

\affil[5]{\orgname{School of Public Health of Sun Yat-sen University}, \orgaddress{\city{Guangzhou}, \postcode{510080}, \state{Guangdong}, \country{China}}}


\abstract{
Large Language Models (LLMs) have demonstrated considerable potential in general practice. However, existing benchmarks and evaluation frameworks primarily depend on exam-style or simplified question-answer formats, lacking a competency-based structure aligned with the real-world clinical responsibilities encountered in general practice. Consequently, the extent to which LLMs can reliably fulfill the duties of general practitioners (GPs) remains uncertain. In this work, we propose a novel evaluation framework to assess the capability of LLMs to function as GPs. Based on this framework, we introduce a general practice benchmark (GPBench), whose data are meticulously annotated by domain experts in accordance with routine clinical practice standards. We evaluate ten state-of-the-art LLMs and analyze their competencies. 
Our findings indicate that current LLMs are not suitable for autonomous deployment in clinical general practice and that all realistic applications require continuous human oversight; further optimization specifically tailored to the daily responsibilities of GPs remains essential.}

\keywords{General Practitioners, Large Language Models, Competency Model, Benchmark}



\maketitle

\section{Introduction}\label{introduction}
Large Language Models (LLMs) have emerged as a prominent technology in recent years, demonstrating remarkable advancements in mathematical reasoning, drug discovery, chemical property prediction, and various other domains. Owing to their extensive knowledge bases and sophisticated reasoning capabilities, LLMs hold substantial potential for applications in the healthcare sector~\cite{DH1, DH2}. 
MedPaLM~\cite{med-palm} and MedPaLM 2~\cite{med-palm2} have reported performance on specific medical tasks that rivals or even surpasses that of human experts. GPT-4 outperformed medical students in open-ended clinical reasoning examinations~\cite{strong2023chatbot}. ChatGPT demonstrated a performance level comparable to that of a third-year medical student on evaluations of core medical knowledge~\cite{gilson2023does}. Med-PaLM~\cite{med-palm} was the first model to attain a passing score on the MedQA dataset~\cite{medqa}, which consists of questions modeled after the U.S. Medical Licensing Examination (USMLE). AMIE~\cite{googleAMIE2025} has also demonstrated effectiveness in assisting physicians with generating potential disease diagnoses.

To evaluate whether LLMs can serve as general practitioners (GPs), it is essential to consider the distinct characteristics of this clinical domain. General practitioners follow a unique framework of clinical thinking: it emphasizes symptom-oriented differential diagnosis rather than definitive diagnosis, considers the interplay among multiple physiological systems rather than focusing solely on organ-specific pathologies, and requires a holistic application of the biopsychosocial model.
Moreover, given the context in which LLMs serve as GPs, their responses should include accurate disease identification and comprehensive diagnoses, offer personalized treatments with clearly defined targets, and incorporate relevant health management considerations.

Most existing benchmarks for LLMs still rely on exam-style formats~\cite{xiaolan2025evaluating,sarraju2023appropriateness,zhu2023can}. Although recent studies have introduced more clinically oriented evaluations~\cite{aime,johri2025evaluation}, these efforts remain largely confined to specialist-level tasks and thus fail to reflect the open-ended reasoning and decision-making integral to real-world general practice, where physicians must ensure continuity of care, triage under uncertainty, and manage multimorbidity in community settings.

\begin{figure}[htbp]
  \centering
  \includegraphics[width=\textwidth]{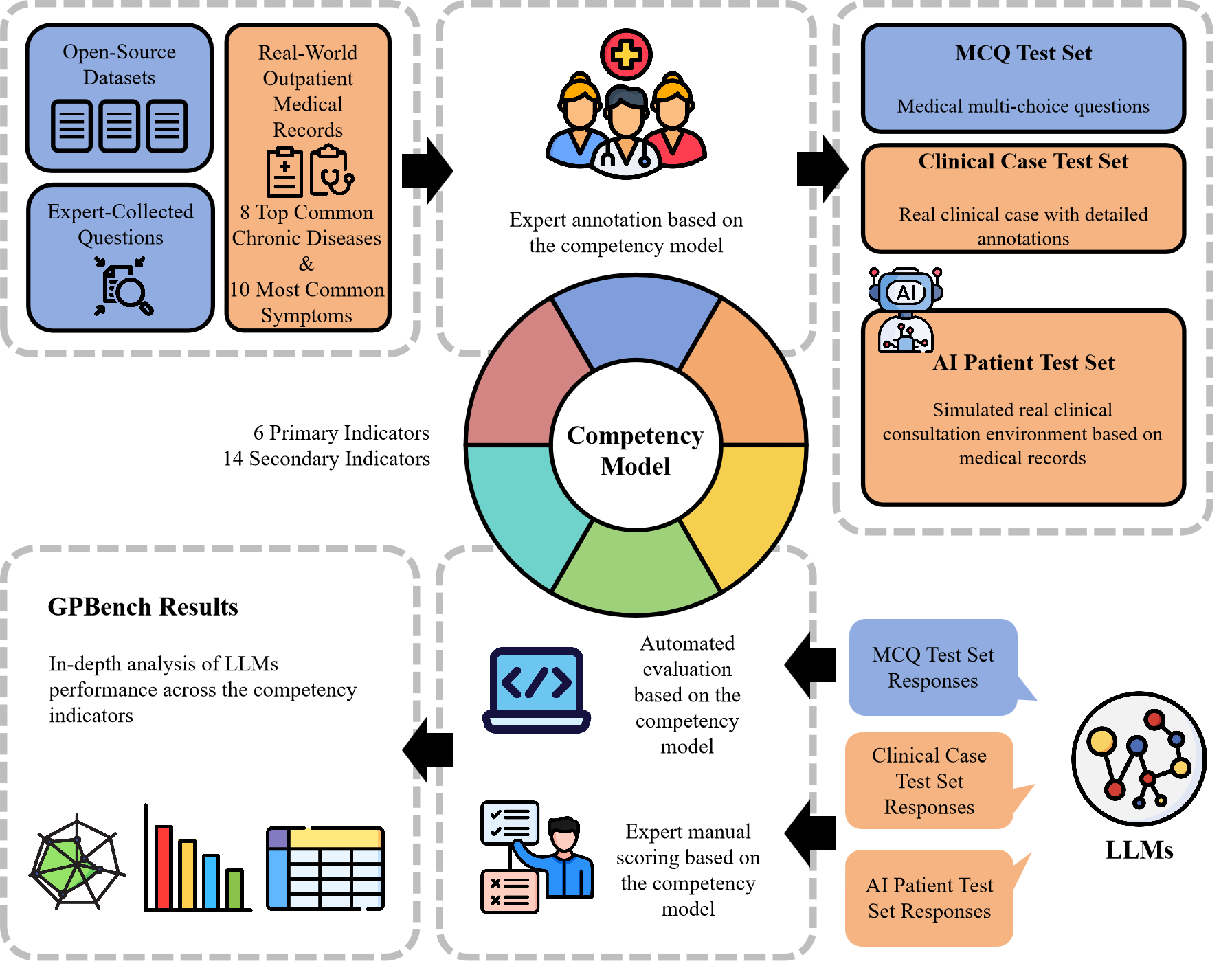}
  \caption{An overview of GPBench.
GPBench comprises a competency model and an evaluation dataset. The competency model delineates six primary indicators and fourteen secondary indicators, capturing the core competencies essential for routine general practice.
Based on the competency model, we collected data from open-source datasets and real outpatient medical records from Tertiary A-grade hospitals to create an evaluation dataset, which contains three parts: the Multiple-choice Question (MCQ) Test Set,  Clinical Case Test Set, and  Artificial Intelligence (AI) Patient Test Set.
Ground truth and scoring criteria for each case were annotated in detail by experts. Accuracy is used as the evaluation metric for the MCQ Test Set, while for the other two test sets, a three-member general practitioner panel grades responses based on the annotated scoring criteria.
}
\label{fig:overview}
\end{figure}

\begin{figure}[htbp]
\includegraphics[width=0.65\textwidth]{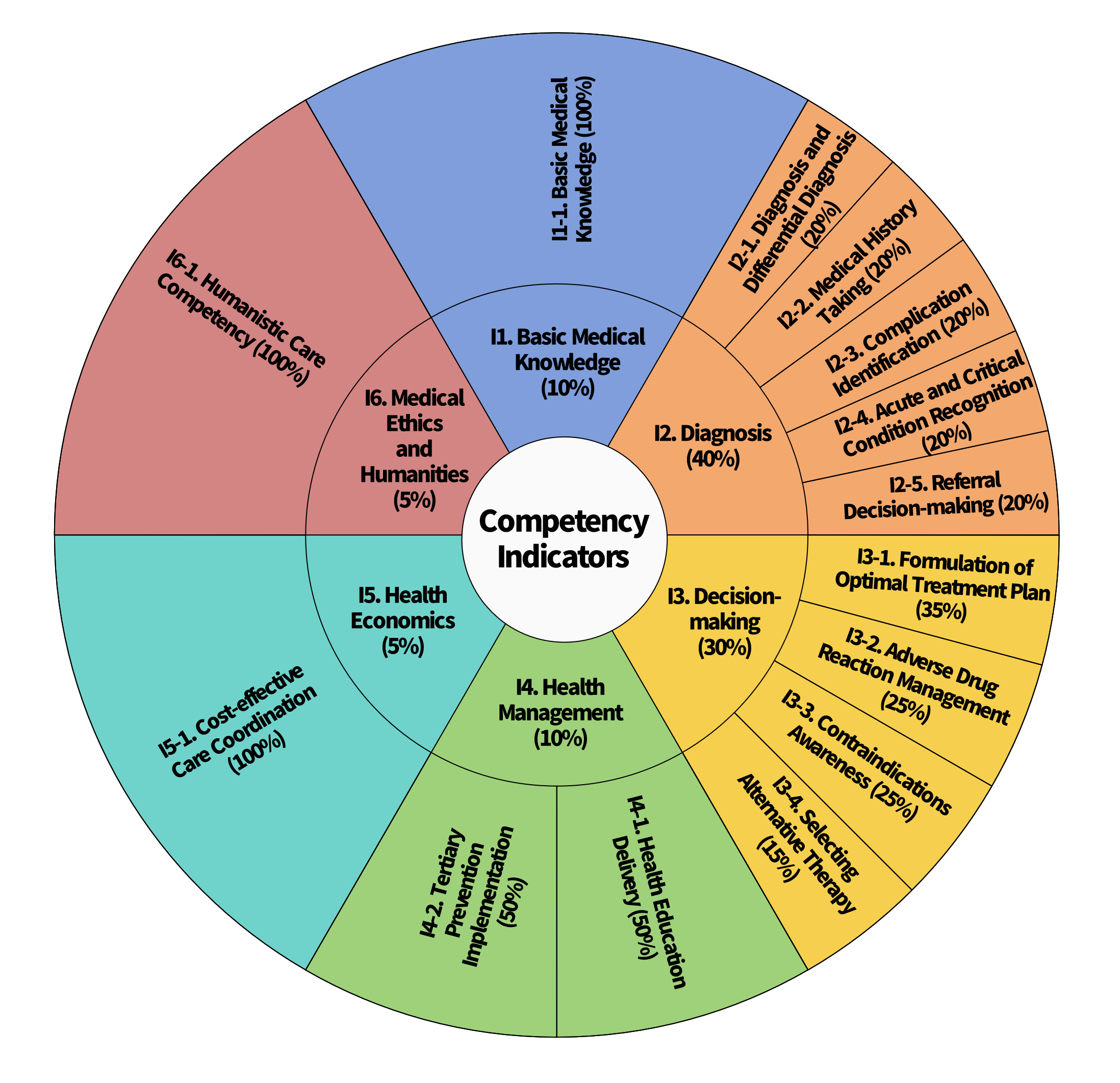}
\centering
\caption{\fontsize{10pt}{10pt}\selectfont An overview of the competency indicators and their associated importance weights in our evaluation framework.}
\label{fig:competency_model}
\end{figure}

To address the aforementioned issues, we propose GPBench to evaluate LLMs in a manner analogous to the evaluation of GPs. As illustrated in Fig.~\ref{fig:overview}, GPBench comprises a competency model and an evaluation dataset.
The competency model, detailed in Fig.~\ref{fig:competency_model}, was derived from real-world clinical practice and delineates the core competencies and decision-making processes essential to routine general practice. It defines six primary indicators—basic medical knowledge, diagnosis, decision-making, health management, health economics, and medical ethics and humanities—and fourteen secondary indicators that specify measurable capabilities under each primary category.
The evaluation dataset consists of three components: a Multiple-choice Question (MCQ) Test Set, a Clinical Case Test Set, and an Artificial Intelligence (AI) Patient Test Set. Additionally, we constructed a separate development set, which remained completely independent from the evaluation dataset and was excluded from all testing procedures.
All evaluation tasks in this study were conducted in Chinese to ensure consistency with the working language of both physicians and annotators.

\begin{table}[h]
\caption{The information of large language models (LLMs) evaluated in our experiments.``Multilingual" indicates models capable of understanding multiple languages, including Chinese.}
\label{tab:evaluated_llms}
\fontsize{8pt}{10pt}\selectfont
\begin{tabular}{>{\centering\arraybackslash}m{0.25\linewidth} 
                >{\centering\arraybackslash}m{0.08\linewidth} 
                >{\centering\arraybackslash}m{0.16\linewidth} 
                >{\centering\arraybackslash}m{0.15\linewidth} 
                >{\centering\arraybackslash}m{0.1\linewidth} 
                >{\centering\arraybackslash}m{0.1\linewidth}}
\toprule
\multirow{2}{*}[-8pt]{Model} &
\multirow{2}{*}[-8pt]{Parameter} &
\multirow{2}{*}[-8pt]{\makecell[c]{Chinese Corpus\\Proportion}}&
\multirow{2}{*}[-8pt]{\makecell[c]{Supported\\Languages}}  &
\multicolumn{2}{c}{Type} \\
\cmidrule{5-6}
 &  &  &  & Medical Specialist & Reasoning Model \\
\midrule
GPT-4o~\cite{hurst2024gpt4o} & -  & Not Reported & Multilingual  & No    & No \\
GPT-4-turbo & - & Not Reported & Multilingual   & No    & No \\
o1-preview~\cite{openaio1}  & -  & Not Reported & Multilingual   & No    & Yes \\
Gemini-1.5-pro~\cite{gemini1.5} & -  & Not Reported & Multilingual  & No    & No \\
Qwen2.5-7B-Instruct~\cite{qwen2.5} & $7$B  & Not Reported & Multilingual  & No    & No \\
Qwen2.5-72B-Instruct~\cite{qwen2.5} & $72$B & Not Reported & Multilingual  & No    & No \\
Claude-3.5-Sonnet & -  & Not Reported & Multilingual   & No    & No \\
DeepSeek-V3~\cite{deepseekv3} & $671$B & Not Reported & Multilingual & No    & No \\
DeepSeek-R1~\cite{deepseekr1} & $671$B & Not Reported & Multilingual & No    & Yes \\
HuatuoGPT-o1-7B~\cite{huatuogpto1} & $7$B & Not Reported & Chinese/English    & Yes   & Yes           \\ 
\bottomrule
\end{tabular}
\end{table}

\section{Results}\label{results}
We used GPBench to evaluate representative models, as shown in Table~\ref{tab:evaluated_llms}, including general-purpose models, reasoning models, and medical-specialist models. A reasoning model referred to a type of LLM specifically designed or trained to emulate human-like reasoning processes, thereby exhibiting enhanced reasoning capabilities. A medical-specialist model was an LLM fine-tuned using carefully curated data from the medical domain.
The evaluation experiments were conducted using the default hyperparameters available on the official release websites or those recommended by the respective providers (see Supplementary Table~\ref{tab:detail_parameters} for details).
We also conducted a human evaluation involving general practitioners to provide a reference for interpreting the performance of LLMs.

\subsection{Results on the Three Test Sets}
The results of reasoning models, general models, and medical specialists on GPBench are discussed based on the indicators defined in the constructed competency model. To enable a more fine-grained capability analysis, some analyses are further conducted at the level of secondary indicators. An overview of all indicators in GPBench is provided in Fig.~\ref{fig:competency_model} (see Supplementary Table~\ref{GPC} for detailed descriptions).

\begin{figure}[t]
\includegraphics[width=1\textwidth]{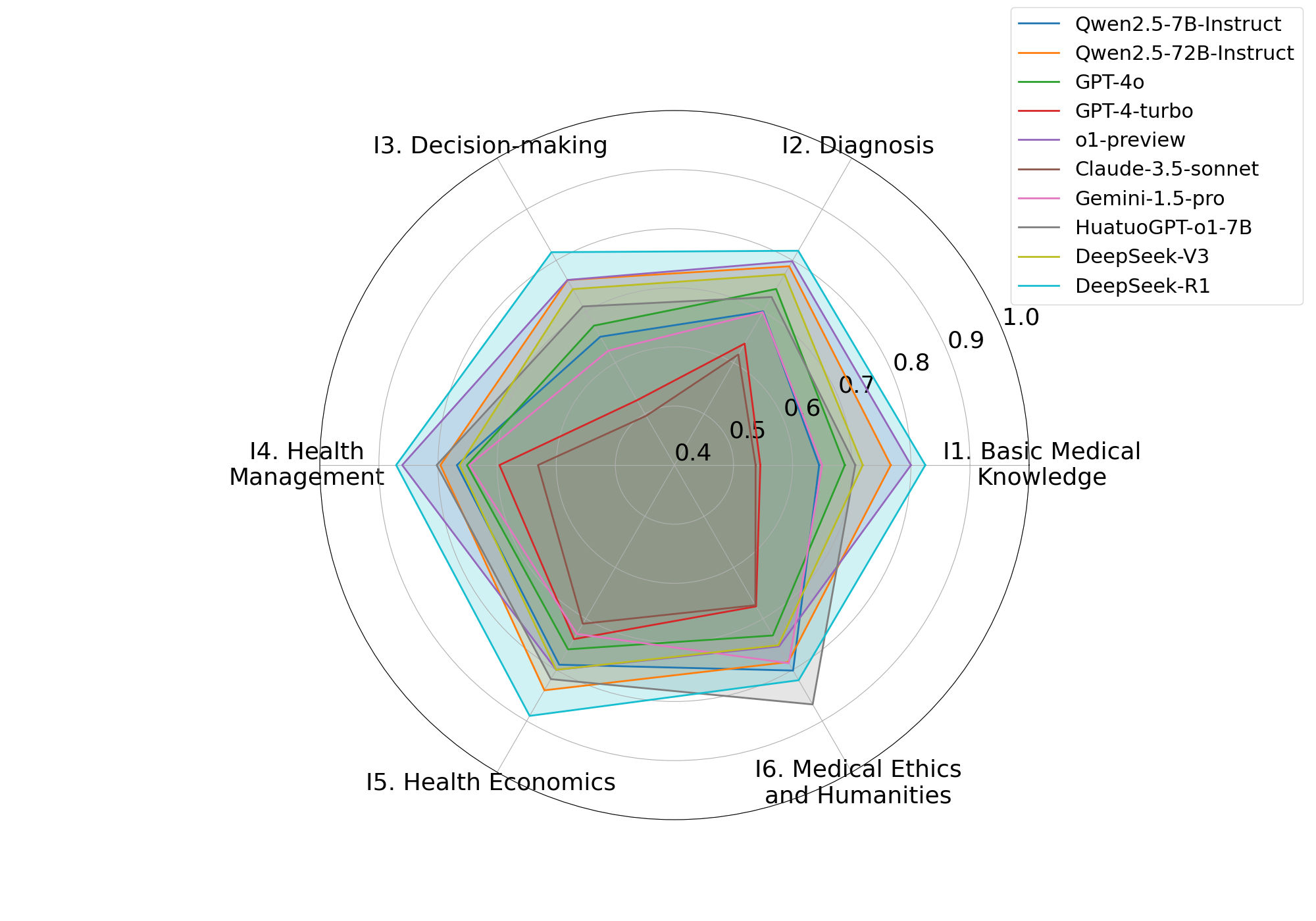}
\centering
\caption{The performance of large language models (LLMs) on the Multiple-choice Question (MCQ) Test Set across the primary competency indicators.}
\label{fig:mcq_result_dim_1}
\end{figure}

\subsubsection{Evaluation Results on the MCQ Test Set}
The evaluation on the MCQ Test Set encompassed 6 primary indicators and 14 secondary indicators. The performance of various LLMs with respect to the primary indicators is shown in Fig.~\ref{fig:mcq_result_dim_1}, where the results for each primary indicator were calculated as the weighted average of the corresponding secondary indicators (see Supplementary Table~\ref{results_test_1} for more details).
Overall, reasoning models, DeepSeek-R1 ($82.74 \pm 1.20$) and o1-preview ($79.16 \pm 1.27$), demonstrated relatively strong performance, albeit with certain limitations. Specifically, DeepSeek-R1 exhibited balanced performance across indicators but did not reach a satisfactory level. In contrast, o1-preview, with accuracy falling below $80$ in several indicators such as I3, I5, and I6, showed varying adaptability to different tasks. This suggested that although models with advanced reasoning capabilities exhibited greater potential, they still faced challenges in fine-grained decision-making tasks. In addition, the overall performance of general-purpose LLMs was relatively weaker than that of reasoning-optimized LLMs, as evidenced by the comparison between their respective top-performing models (Qwen2.5-72B-Instruct: $78.13 \pm 1.28$ vs. DeepSeek-R1: $82.74 \pm 1.20$; Welch's $t$-test, $p = 0.0043$). 
Meanwhile, Qwen2.5-72B-Instruct significantly outperformed Qwen2.5-7B-Instruct ($78.13 \pm 1.28$ vs. $69.63 \pm 1.40$; Welch's $t$-test, $p < 0.0001$), indicating that LLMs with larger parameter scales tended to achieve better performance.

\begin{figure}[!h]
\includegraphics[width=1\textwidth]{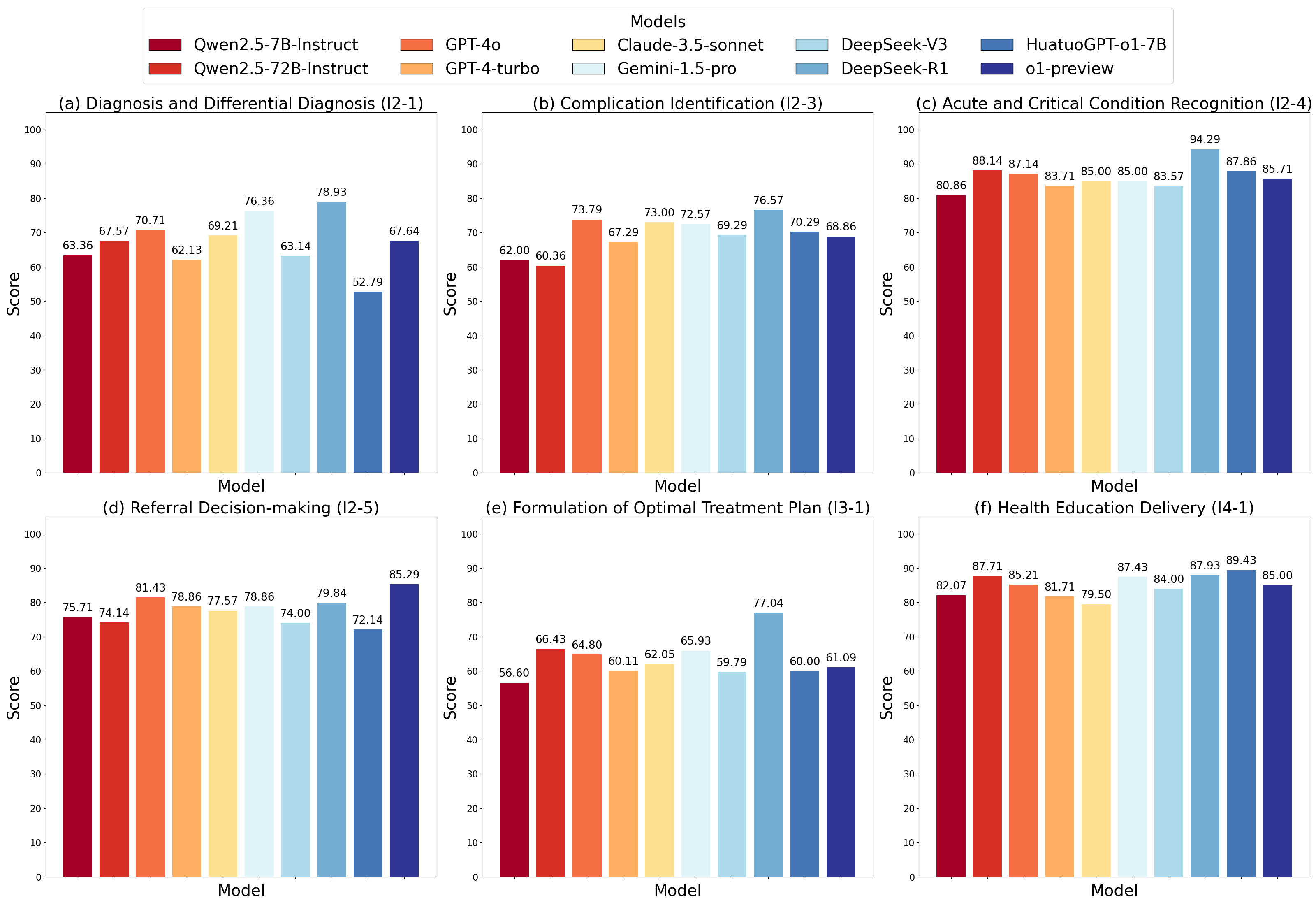}
\centering
\caption{\fontsize{10pt}{10pt}\selectfont The performance of large language models (LLMs) on the Clinical Case Test Set across multiple competency indicators. 
(a) Performance on diagnosis and differential diagnosis (I2-1). 
(b) Performance on complication identification (I2-3). 
(c) Performance on acute and critical condition recognition (I2-4).
(d) Performance on referral decision-making (I2-5).
(e) Performance on formulation of optimal treatment plan (I3-1).
(f) Performance on health education delivery (I4-1).
}
\label{fig:clinical_case_results_bar}
\end{figure}

The medical specialist HuatuoGPT-o1-7B demonstrated strong performance in I4 ($80.23 \pm 3.81$), I5 ($81.82 \pm 8.22$), and I6 ($86.76 \pm 3.70$), whereas its results in the remaining three primary indicators—I1 ($70.59 \pm 1.10$), I2 ($72.85 \pm 2.30$), and I3 ($70.45 \pm 2.58$)—were at a moderate level.
This disparity suggested that current medical specialist LLMs were fine-tuned without comprehensive optimization across diverse medical competencies.

\subsubsection{Evaluation Results on the Clinical Case Test Set}
For the Clinical Case Test Set, the inputs of LLMs were medical records, and the target outputs were the corresponding diagnosis and treatment.
This test set was specifically designed to reflect the real-world demands of general practice. Representative cases were selected for inclusion, with the aim of covering as many different diseases, symptoms, and levels of difficulty as possible. As a result, each case exhibited significant attribute variation, which in turn affects the performance of LLMs, leading to considerable fluctuation in test results across different cases.
On this test set, some secondary indicators (e.g., I1-1, I2-2, I4-2, I5-1, I6-1) could not be evaluated. Thus, six secondary indicators—those that were manifested in the generations—were selected for the final evaluation. 
Fig.~\ref{fig:clinical_case_results_bar} presents the performance of LLMs across multiple competency indicators (see Supplementary Table~\ref{results_cases} for more details).

We found that most LLMs showed significant shortcomings in the core aspects of competencies. Specifically, in the I2-1 indicator, $7/10$ models scored below $70$, indicating that most LLMs failed to construct a complete chain of diagnostic reasoning. In the I3-1 indicator, only DeepSeek-R1 ($77.04 \pm 21.33$) scored above $70$, indicating that current LLMs were basically unable to generate satisfactory treatment plans. 

DeepSeek-R1 outperformed other models across indicators related to diagnosis and treatment, ranking at the forefront across all indicators (DeepSeek-R1: $81.80 \pm 16.92$ vs. Gemini-1.5-pro: $76.46 \pm 20.14$; Wilcoxon rank-sum test, $p = 0.046$), with Gemini-1.5-pro demonstrating the second-best performance. This statistically significant difference highlighted a potential direction for optimizing LLMs in general practice applications. Furthermore, the results underscored the superiority of reasoning-optimized LLMs over general-purpose LLMs, as DeepSeek-R1 and Gemini-1.5-pro represented the top-performing models within their respective model categories. Meanwhile, Qwen2.5-72B-Instruct outperformed Qwen2.5-7B-Instruct ($72.79 \pm 17.18$ vs. $68.63 \pm 16.26$; Wilcoxon rank-sum test, $p = 0.010$), suggesting that larger models tended to achieve better performance.

Moreover, despite explicit prompts, none of the LLMs proactively demonstrated I5 or I6. This suggested that current LLMs remained limited to a simplified framework of diagnosis and treatment, lacking a broader understanding of the social attributes of healthcare services. 
We further revealed the deficiencies of LLMs through a detailed analysis of the responses, as described in section~\ref{3.2}.

\begin{figure}[!h]
\includegraphics[width=0.55\textwidth]{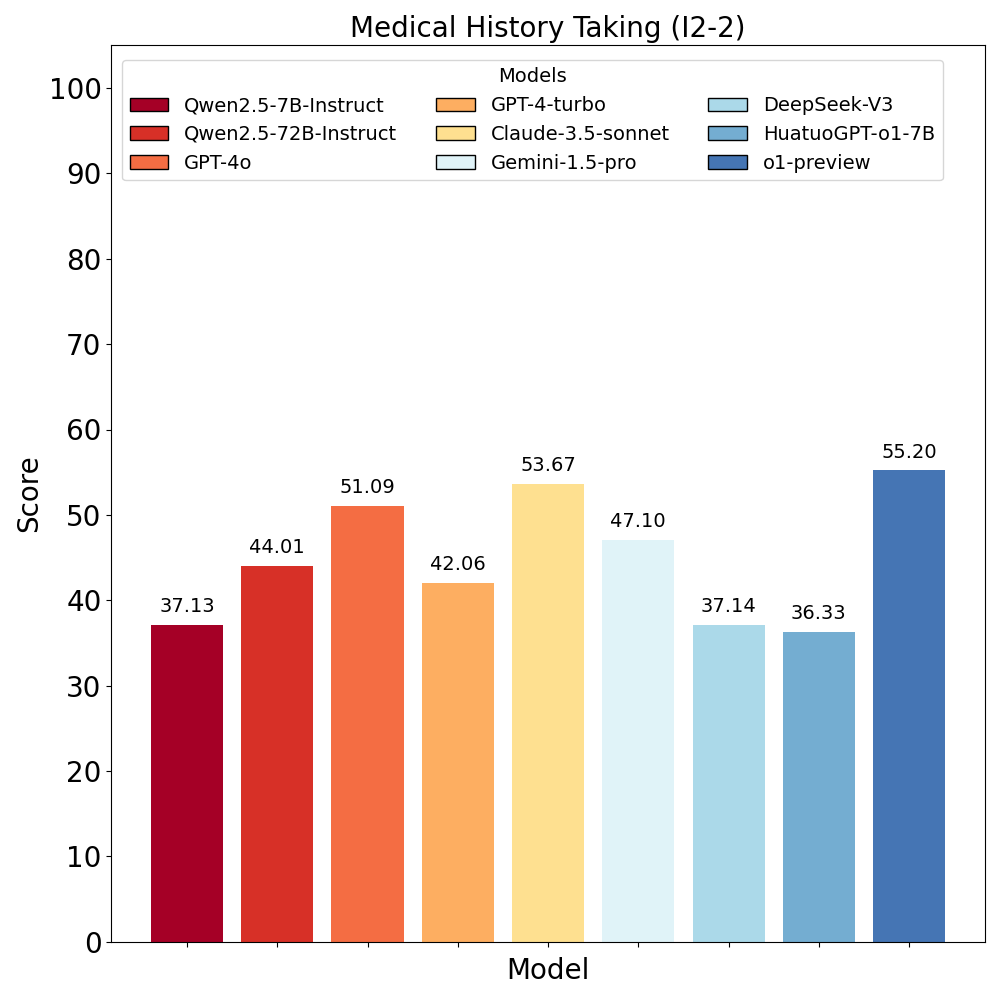}
\centering
\caption{\fontsize{10pt}{10pt}\selectfont The performance of large language models (LLMs) on the Artificial Intelligence (AI) Patient Test Set for the Medical History Taking (I2-2) indicator.}
\label{fig:ai_patient_results_bar}
\end{figure}

\subsubsection{Evaluation Results on the AI Patient Test Set}
The AI Patient Test Set evaluated LLMs' competency in medical history taking by measuring their ability to ask relevant questions. Fig.~\ref{fig:ai_patient_results_bar} shows the resulting competency scores. DeepSeek-R1 was excluded due to its substantial failure to follow instructions and meet basic evaluation criteria.

From Fig.~\ref{fig:ai_patient_results_bar}, we found that in the I2-2 indicator, all models scored below $60$, with the best performer, o1-preview, achieving only $55.20 \pm 8.01$. 
LLMs exhibited significant deficiencies in medical history taking, ultimately hindering their ability to support effective clinical decision-making.
In one of the abdominal pain tests, o1-preview showed significant weaknesses. It failed to adequately explore critical pain features such as intensity, radiation, and triggering or relieving factors—essential for distinguishing conditions like appendicitis and cholecystitis. The assessment of associated symptoms was superficial, limited to checking for vomiting and fever, without further inquiry into important diagnostic clues like the nature of the vomitus or the presence of chills. Additional issues included the omission of other potentially relevant symptoms, incomplete collection of personal and medication history, and failure to assess genitourinary, respiratory, and gastrointestinal systems. Past medical and family histories were also insufficiently gathered.

\subsection{Human Performance as a Reference}

To benchmark the performance of LLMs against human physicians, we evaluated GPs using the same test sets. The participating GPs were classified into three groups: Group A ($> 10$ years), Group B ($5 \sim 10$ years), and Group C ($<5$ years). This stratification allowed us to position the capabilities of LLMs within a meaningful spectrum of real-world clinical expertise. The scores of the human evaluation can be found in Supplementary Table~\ref{tab:human_123}.

On the MCQ Test Set, LLMs generally outperformed human participants in terms of accuracy. This outcome was anticipated, as the format restricts possible answers and mainly tests factual knowledge recall, an area where LLMs excel owing to their extensive pretraining on large-scale factual text corpora.

In contrast, on the Clinical Case Test Set, LLMs performed worse than human physicians overall. Among the evaluated models, the best-performing LLM, DeepSeek-R1, reached a competency level comparable to that of Group C human physicians, while the remaining models exhibited inferior performance. Competency analysis revealed that LLMs had substantial limitations in core diagnostic and treatment decision-making capabilities, falling short even when compared to Group C human physicians. Nonetheless, LLMs demonstrated a distinct advantage in health education, outperforming experienced Group A human physicians—likely attributable to their ability to organize and present information in a well-structured manner.

The gap further widened in the medical history taking competency on the AI Patient Test Set, where even the highest-performing model, o1-preview, underperformed compared to Group C human physicians. This limitation likely stems from LLMs’ lack of real clinical experience and insufficient exposure to temporally interactive diagnostic processes, leading to logical discontinuities and incomplete or fragmented histories.

Overall, these results indicate that while LLMs have matched or even surpassed human physicians in assessments of static knowledge, they still lag significantly behind in interactive and reasoning-intensive clinical tasks. Their strengths lie in factual recall and the clear articulation of information, yet they have not attained the practical competency level of even less experienced general practitioners in real-world diagnostic and patient-interview scenarios.

\subsection{Competency Analysis} \label{3.2}
We analyzed LLM responses on the Clinical Cases Test Set using the proposed evaluation framework to identify key deficiencies in core competencies. For diagnosis and treatment recommendations, we reviewed all outputs and categorized common deficiency types. Each case was annotated with the relevant deficiencies per LLM, allowing for multiple deficiencies per case. For each deficiency type, we computed the proportion of errors per LLM across applicable cases. Results are shown in Table~\ref{tab:diagdef} and Table~\ref{tab:treatdef}, respectively.

\begin{table}[h]
\centering
\caption{The proportion of observed deficiencies of large language models (LLMs) in the diagnostic process on the Clinical Case Test Set.}
\label{tab:diagdef}
\fontsize{8pt}{10pt}\selectfont
\begin{tabular}{
  >{\centering\arraybackslash}m{0.12\linewidth}
| >{\centering\arraybackslash}m{0.12\linewidth}
 >{\centering\arraybackslash}m{0.12\linewidth}
 >{\centering\arraybackslash}m{0.12\linewidth}
>{\centering\arraybackslash}m{0.12\linewidth}
 >{\centering\arraybackslash}m{0.12\linewidth}}
\hline
Model & Lack of Accurate Disease Grading, Staging, and Risk Stratification (Applicable cases $N_1 = 27$) & Hallucination on Disease Grading, Staging, and Risk Stratification (Applicable cases $N_2 = 43$) & Blind Spots in Identification of Comorbidities and Complications (Applicable cases $N_3 = 45$) & Deficiency in Acute and Critical Condition Assessment (Applicable cases $N_4 = 18$) & Lack of Accurate Diagnosis for Rare Diseases (Applicable cases $N_5 = 3$)\\ 
\hline
Qwen2.5-7B-Instruct & $20/27$ & $16/43$ & $23/45$ & $3/18$ & $2/3$\\
Qwen2.5-72B-Instruct & $14/27$ & $42/43$ & $13/45$ & $1/18$ & $1/3$\\
GPT-4o & $\ 9/27$ & $11/43$ & $26/45$ & $2/18$ & $2/3$\\
GPT-4-turbo & $13/27$ & $6/43$ & $25/45$ & $3/18$ & $3/3$\\
Claude-3.5-Sonnet & $11/27$ & $17/43$ & $26/45$ & $3/18$ & $2/3$\\
Gemini-1.5-pro & $11/27$ & $14/43$ & $19/45$ & $2/18$ & $1/3$\\
DeepSeek-V3 & $15/27$ & $15/43$ & $22/45$ & $3/18$ & $2/3$\\
DeepSeek-R1 & $7/27$ & $18/43$ & $8/45$ & $4/18$ & $2/3$\\
HuatuoGPT-o1-7B & $12/27$ & $1/43$ & $31/45$ & $2/18$ & $2/3$\\
o1-preview & $15/27$ & $18/43$ & $17/45$ & $2/18$ & $1/3$ \\ \hline
\end{tabular}
\end{table}

\subsubsection{Inadequate Diagnostic Competency}

We conducted a detailed examination of LLM-generated diagnostic outputs and identified common deficiencies in diagnostic competency. Our analysis was presented as follows.

(1) Lack of accurate disease grading, staging, and risk stratification.
LLMs tended to oversimplify the management of diseases that required classification or risk stratification. As shown in Table~\ref{tab:diagdef}, such a deficiency was common in LLMs with Qwen2.5-7B-Instruct showing the highest proportion at $20/27$. In the case tests for HTN, CAD, and CKD, most LLMs, including Qwen2.5-72B-Instruct, o1-preview, and DeepSeek-R1, failed to perform grading, staging, and risk stratification according to the corresponding specialty guidelines. 
Specifically, in one of the CAD case tests, $7/10$ models only provided a general diagnosis of ``coronary artery disease/stable angina'' without specifying cardiac function classification.
Moreover, some models even exhibited incorrect disease grading and staging. For example, in one of the CKD case tests, Qwen2.5-72B-Instruct and GPT-4o misdiagnosed CKD stage $5$ as CKD stage $3$.

(2) Hallucination on disease grading, staging, and risk stratification.
LLMs arbitrarily graded and staged diseases without clinical guidelines or evidence-based medical support, leading to a significant issue of fabricating clinical staging during diagnosis, which involved a variety of diseases such as pneumococcal pneumonia, diabetes, gouty arthritis, osteoarthritis, pneumothorax, and acute appendicitis. 
For example, in one of the abdominal pain case tests, Qwen2.5-72B-Instruct provided the diagnoses of ``acute appendicitis Grade 0''.
As shown in Table~\ref{tab:diagdef}, Qwen2.5-72B-Instruct exhibited the highest proportion $42/43$, with DeepSeek-R1 also reaching $18/43$. In contrast, HuatuoGPT-o1-7B demonstrated a much lower proportion of only $1/43$, suggesting that LLMs that had undergone domain-specific fine-tuning were considerably less susceptible to hallucinations.

(3) Blind spots in identification of comorbidities and complications.
Current LLMs exhibited insufficient awareness in systematically screening for comorbidities and complications, especially in the scenarios of metabolic and cardiovascular diseases, where key diagnostic elements were frequently omitted, 
with a deficiency proportion approximately approaching $50\%$ among applicable cases involving this deficiency ($N_3 =45$) for LLMs shown in Table~\ref{tab:diagdef}. This issue was mainly manifested in two aspects: 
Failed to fully identify comorbidities and complications. For example, in one of the HLD case tests, almost all models failed to diagnose additional comorbidities or complications, with the absence of a diagnosis of severe fatty liver and HLD.
Incorrectly assess comorbidities and complications. For instance, in one of the HLD case tests,  Qwen2.5-72B-Instruct and Gemini-1.5-pro incorrectly diagnosed diabetic nephropathy (DN), although the patient did not meet the diagnostic criteria for DN. 

(4) Deficiency in acute and critical condition assessment.
LLMs showed significant shortcomings in the identification of clinical emergencies, particularly in the recognition of time-sensitive acute and life-threatening critical conditions, with a deficiency proportion of approximately $10\%$ among the applicable cases ($N_4 =18$) according to Table~\ref{tab:diagdef}.

(5) Lack of accurate diagnosis for rare diseases.
LLMs exhibited limitations in diagnosing rare diseases in our evaluations, as rare diseases were often associated with complex pathological mechanisms and heterogeneous clinical manifestations. For instance, in one of the fever case tests, $8/10$ LLMs failed to provide an accurate diagnosis for the rare disease scrub typhus.

\begin{table}
\caption{The proportion of observed deficiencies of large language models (LLMs) in treatment recommendations on the Clinical Case Test Set.}\label{tab:treatdef}
\fontsize{8pt}{10pt}\selectfont
\begin{tabular}{>{\centering\arraybackslash}m{0.09\linewidth}
               |>{\centering\arraybackslash}m{0.10\linewidth}
               >{\centering\arraybackslash}m{0.07\linewidth}
               >{\centering\arraybackslash}m{0.07\linewidth}
               >{\centering\arraybackslash}m{0.07\linewidth}
               >{\centering\arraybackslash}m{0.08\linewidth}
               >{\centering\arraybackslash}m{0.11\linewidth}
               >{\centering\arraybackslash}m{0.13\linewidth}}
\hline
\multirow{2}{=}[-36pt]{Models} & \multirow{2}{=}[-10pt]{Lack of Treatment Goals (Applicable cases $N_6 = 28$)} & \multicolumn{4}{c}{\makecell[c]{Pharmacotherapy Risks \\ (Applicable cases $N_7 = 70$)}} & \multirow{2}{=}{Blind Spots in Non-Pharmacological Interventions (Applicable cases $N_8 = 70$)} & \multirow{2}{=}{Insufficient Standardization in Critical Illness Management (Applicable cases $N_9 = 18$)} \\
\hhline{|~|~|----|~|~|}
 &  & Omission of Core Drugs & Neglect of Drug Interaction Contraindications & Lack of Drug Dosage Guidance & Improper control of medication indications &  & \\
\hline
Qwen2.5-7B-Instruct & $17/28$ & $17/70$ & $1/70$ & $12/70$ & $6/70$ & $14/70$ & $9/18$\\
Qwen2.5-72B-Instruct & $16/28$ & $18/70$ & $1/70$ & $7/70$ & $3/70$ & $9/70$ & $8/18$\\
GPT-4o & $14/28$ & $18/70$ & $4/70$ & $6/70$ & $2/70$ & $11/70$ & $8/18$\\
GPT-4-turbo & $17/28$ & $21/70$ & $2/70$ & $17/70$ & $2/70$ & $11/70$ & $9/18$\\
Claude-3.5-Sonnet & $18/28$ & $24/70$ & $2/70$ & $12/70$ & $2/70$ & $10/70$ & $8/18$\\
Gemini-1.5-pro & $11/28$ & $16/70$ & $1/70$ & $10/70$ & $0/70$ & $11/70$ & $7/18$\\
DeepSeek-V3 & $18/28$ & $20/70$ & $5/70$ & $5/70$ & $1/70$ & $12/70$ & $8/18$\\
DeepSeek-R1 & $12/28$ & $5/70$ & $2/70$ & $2/70$ & $1/70$ & $12/70$ & $2/18$\\
HuatuoGPT-o1-7B & $17/28$ & $20/70$ & $0/70$ & $10/70$ & $3/70$ & $10/70$ & $6/18$\\
o1-preview & $15/28$ & $13/70$ & $2/70$ & $4/70$ & $0/70$ & $10/70$ & $5/18$\\
\hline
\end{tabular}
\end{table}

\subsubsection{Inadequate Treatment Decision-Making Competency}
We examined all the details of outputs generated by the LLMs and identified common deficiencies in treatment decision-making, with the results shown in Table~\ref{tab:treatdef}. Our analysis was presented as follows.

(1) Lack of treatment goals.
In the treatment recommendations generated by LLMs for chronic diseases (e.g.,  HTN and DM), there was a widespread lack of clearly defined therapeutic targets. 
As shown in Table~\ref{tab:treatdef}, this deficiency type accounted for approximately half of the applicable cases ($N_6 = 28$).
For instance, nearly all models failed to explicitly specify control targets for blood pressure or blood glucose in their treatment plans.

(2) Pharmacotherapy risks.
Through case analysis, we found that LLMs exhibited issues related to pharmacotherapy risks, with approximately $40\%$ of the applicable cases ($N_7=70$) demonstrating this deficiency type according to Table~\ref{tab:treatdef}. We described four types of such deficiencies as follows.
\begin{itemize}
    \item Omission of core drugs. 
This was the most common type of pharmacotherapy risk.
Most LLMs omitted the main drugs in the specific disease recommendations, such as CAD, HLD, and CKD. For example, in one of the HLD case tests, $9/10$ models failed to include somatostatin and proton pump inhibitors (PPIs) in the treatment plan for severe acute pancreatitis. 

    \item  Neglect of drug interaction contraindications.
Some LLMs failed to comprehensively evaluate the overall safety of drug combinations, ignoring the potential for increased side effects or adverse reactions.
For example, in one of the HTN case tests, Qwen2.5-7B-Instruct recommended the concurrent use of fibrate and statin drugs (fenofibrate and atorvastatin) for lipid-lowering. 

    \item Lack of drug dosage guidance.
Some LLMs showed insufficient or erroneous guidance on drug dosing.
For example, in one of the HLD case tests, GPT-4-turbo failed to specify the dosages for lipid-lowering drugs and antibiotics. 
In the same case, Qwen2.5-72B-Instruct recommended a dose of Ulinastatin at $10,000 \ \text{U}$, which significantly deviated from the $100,000 \ \text{U}$ recommended by the guidelines.
Moreover, in some cases, certain models even suggested dosing routes that did not match the acute phase of the disease.

    \item Improper control of medication indications.
Some LLMs failed to adequately consider individual patient characteristics and specific disease requirements when providing treatment recommendations.
For example, in one of the CKD case tests, LLMs, including Qwen2.5-7B-Instrcut, Qwen2.5-72B-Instrcut, and Claude-3.5-Sonnet, did not account for the patient's renal function when selecting and adjusting antihyperglycemic drugs.
    
\end{itemize}

(3) Blind spots in non-pharmacological interventions.
We found that the treatment recommendations of LLMs for non-pharmacological treatment plans, such as interventional therapies and surgical indications, were significantly underrepresented.
As shown in Table~\ref{tab:treatdef}, the deficiency type accounted for approximately $15\%$ among the applicable cases ($N_8 = 70$).
For instance, in the case tests for CAD and CKD, LLMs, such as Qwen2.5-72B-Instruct, GPT-4o, and Gemini-1.5-pro, did not provide interventional therapies, dialysis plans, and other options tailored to the disease progression stage of patients.

(4) Insufficient standardization in critical illness management.
In the management recommendations for acute and critical conditions, LLMs failed to adhere to clinical guidelines and even omitted critical steps in certain disease scenarios, with the highest proportion of $9/18$ for Qwen2.5-7B-Instruct and GPT-4-turbo.

\subsubsection{Inadequate Health Management and Humanistic Care Competencies}
We examined the model-generated health management advice and found that current LLMs tended to adopt an overly principled approach when providing health education.
Although the content generally met basic scoring standards, it often remained at a broad, principle-based level, lacking concrete, actionable guidance that would effectively support clinical practice needs. For instance, in the case tests for HTN, most LLMs merely suggested a ``sodium-restricted diet'' without specifying precise daily sodium intake limits (e.g., $\leq 5g$).

\subsubsection{Medical Ethical Risks}

We examined the model-generated outputs and identified the presence of medical ethical risks, which were primarily manifested as an over-treatment tendency and an increased risk of misdiagnosis or missed diagnosis.

(1) Over-treatment tendency.
Some LLMs tended to generate unreasonable treatment recommendations due to algorithmic bias or limitations in training data. This was particularly evident in specific disease scenarios. For example, in one of the HTN case tests, Qwen2.5-7B-Instruct recommended combination therapy with two antihypertensive drugs and two lipid-lowering medications, which was unnecessary for the current patient and inconsistent with clinical guidelines. 

(2) Risk of misdiagnosis or missed diagnosis.
In one of the HTN case tests, the patient was experiencing a hypertensive emergency, but some models misdiagnosed the condition as mild hypertension or an unrelated disorder, such as Qwen2.5-7B-Instruct diagnoses hypertension, GPT-4-turbo diagnoses transient ischemic attack (TIA).

\section{Discussion}\label{sec12}

The evaluation revealed that LLMs generally outperformed human physicians on knowledge-intensive tasks but underperformed in reasoning and interactive competencies critical to general practice. Specifically, LLMs faced challenges in diagnosing, managing multimorbidity, and making treatment decisions, including overlooking key therapeutic objectives, deviating from standard protocols, and failing to construct coherent reasoning frameworks during history-taking, which also raised ethical concerns such as overtreatment or misdiagnosis.
In contrast, human physicians showed gradual improvement with experience, with experienced general practitioners demonstrating more coherent reasoning and reliable clinical judgment.

Potential causes of the deficiencies identified in our study include the following aspects. (1) The hallucination problem inherent in large models. Hallucinations refer to outputs that appear plausible or logical but are, in fact, incorrect, fabricated, or inconsistent with established facts. This issue primarily stems from the fact that LLMs generate responses based on linguistic statistics rather than knowledge-based logic.
(2) The lack of modeling of human physiology, disease progression, and pharmacology. LLMs are trained via next-token prediction over large textual corpora, and therefore lack true capabilities in modeling human physiological and pathophysiological processes. As a result, they often struggle to accurately infer disease trajectories or assess comorbid conditions (e.g., drug interactions in polypharmacy).
(3) Deficiencies in fine-grained information retention and reasoning. This limitation likely stems from the scarcity of structured, case-level data derived from human clinical training and workflows in the models’ training corpora.

Based on the preceding analysis, we propose the following recommendations for the application of LLMs in general practice:

\begin{itemize}
    
    \item LLMs should only be used under the supervision of medical personnel capable of evaluating the correctness of the output.
    \item Integrate local clinical guidelines, case studies, and textbooks to support evidence-based medicine, thereby reducing hallucinations and improving factual accuracy, potentially combined with automated fact-checking mechanisms~\cite{augenstein2024factuality,ng2025rag}.
    \item Prefer LLMs fine-tuned for medical reasoning, with sufficient parameter scale, and consider interactive or multi-agent architectures for complex decision-making~\cite{tran2025multi,kim2024mdagents}.
    \item For languages with limited pretraining data, fine-tune LLMs on high-quality local corpora to enhance reliability and reasoning performance. 

\end{itemize}
 
We have shown the limitations of LLMs in real-world general practitioner scenarios and now discuss the main limitations of our study. 
First, all cases and prompts were in Chinese and collected in China, which may introduce language-related biases. The models' performance in other languages and regions requires further validation. 
Second, we used standard prompting, the most common approach for applying LLMs. Performance may improve with advanced methods, such as interactive multi-agent systems~\cite{multi_agent_npj} and reinforcement learning tailored to medicine~\cite{aime}, though these are beyond the scope of this study, which focuses on evaluating LLMs' native medical capabilities. 
Lastly, our dataset includes samples covering 8 diseases and 10 symptoms, as all samples require fine-grained human annotation. To better reflect the nature of GPs' daily tasks, we plan to develop automated scoring methods to reduce reliance on expert annotation and expand disease coverage.

In conclusion, by leveraging the proposed evaluation method and our fine-grained annotated dataset, we evaluated the performance of current state-of-the-art large models, encompassing both general-purpose and medical-specialized variants. The experimental results indicate that the models assessed in this study still exhibit significant limitations in critical tasks such as clinical decision support, diagnostic accuracy, and treatment recommendation. To improve the performance of large models in real-world general practice scenarios, we propose that future research should focus on the following directions:

\begin{itemize}
\item Further enhancing the systematic medical knowledge (i.e., diseases, drugs, symptoms) and complex reasoning capabilities of large models, while mitigating the occurrence of medical hallucinations in LLMs~\cite{kim2025medical}.
\item Progressively learn from physicians' clinical experience by incorporating structured data derived from clinical workflows into the training process.
\item Strengthening capabilities in human body modeling, disease progression modeling, and pharmacological modeling~\cite{holt2024data}.

\end{itemize}

Our GPBench can effectively reflect the limitations of applying LLMs in general practice and provide valuable guidance for enhancing their capabilities. The dataset will be publicly released to support broad adoption. Through this initiative, we aim to promote the development of LLMs for general practice and contribute to the advancement and optimization of clinical decision support systems.

\section{Methods}\label{sec11}
\bmhead{Ethics approval and consent to participate}
This study was approved by the Institutional Review Boards of The Sixth Affiliated Hospital of Sun Yat-sen University (IRB No.2025ZSLYEC-117) and Xinyi People’s Hospital (IRB No.CT-2025-03-01). The requirement for informed consent was waived due to the anonymized nature of the data and minimal risk to participants. All patient identifiers were removed prior to analysis. 
Data governance and anonymization procedures were conducted under the oversight of the participating hospitals’ institutional data protection and ethics governance frameworks, in compliance with the General Data Protection Regulation (GDPR).
The authors confirm no conflicts of interest related to this work.

\subsection{Evaluation Framework} 
To evaluate LLMs in a manner analogous to the assessment of GPs, we draw upon widely accepted competency frameworks in general practice, including WONCA~\cite{wonca2012definition}, ACGME~\cite{scherger2007preparing}, 
the Iceberg Model~\cite{mcclelland1973testing}, and the Onion Model~\cite{boyatzis1991competent},
and established an evaluation framework using the Delphi expert consensus method. We formed an expert committee composed of ten general medicine experts and two medical informatics experts (see Supplementary Table~\ref{tab:annotators}).
The committee conducted a comprehensive analysis of the core competencies required for GPs in daily practice and selected indicators that possess both technical characteristics and clinical applicability, grounded in the practical deployment of LLMs in general practice.
Competencies in non-technical domains, such as organizational management, leadership, and teamwork, were excluded. After three rounds of iterative consultation, the committee proposed a competency model specifically tailored for LLM evaluation.
The resulting competency model for LLMs comprised six primary general practice competency indicators and fourteen secondary indicators, each associated with a corresponding importance weight. The complete competency model is illustrated in Fig.~\ref{fig:competency_model} (see Supplementary Table~\ref{GPC} for a detailed description of each indicator). It encapsulated the core competencies essential for routine clinical activities of GPs, which also aligned with the expected capabilities of LLMs in real-world scenarios.

\begin{table}
\caption{Overview of the proposed benchmark.}\label{tab:gpbench_overview}
\centering
\fontsize{8pt}{10pt}\selectfont
\begin{tabular}{ >{\centering\arraybackslash}m{0.15\linewidth}
                    |>{\centering\arraybackslash}m{0.50\linewidth}
                    |>{\centering\arraybackslash}m{0.13\linewidth}
                    |>{\centering\arraybackslash}m{0.11\linewidth}}

\hline
Test Set & Content & Format & Number\\ \hline
MCQ Test Set & Open-source data and expert supplementation  & Objective questions & $3,661$\\ \hline
\multirow{12}{=}{Clinical Case Test Set} & \multirow{24}{=}{\makecell[l]{
Outpatient medical records containing any of \\ the following 8 major chronic diseases or 10 \\ common symptoms: \\
\labelitemi\hspace{\dimexpr\labelsep+0.5\tabcolsep}8 Major Chronic Diseases: \\ \hspace*
{0.5\leftmargin}\labelitemii\hspace{\dimexpr\labelsep+0.5\tabcolsep}Hypertension (HTN)\\\hspace*{0.5\leftmargin}\labelitemii\hspace{\dimexpr\labelsep+0.5\tabcolsep}Hyperlipidemia (HLD) \\\hspace*{0.5\leftmargin}\labelitemii\hspace{\dimexpr\labelsep+0.5\tabcolsep}Coronary Artery Disease (CAD)\\\hspace*{0.5\leftmargin}\labelitemii\hspace{\dimexpr\labelsep+0.5\tabcolsep}Chronic Kidney Disease (CKD)\\\hspace*{0.5\leftmargin}\labelitemii\hspace{\dimexpr\labelsep+0.5\tabcolsep}Chronic Obstructive Pulmonary\\\hspace*{0.5\leftmargin}\hspace*{\dimexpr\labelsep+1.3\tabcolsep}Disease (COPD)\\\hspace*{0.5\leftmargin}\labelitemii\hspace{\dimexpr\labelsep+0.5\tabcolsep}Cerebrovascular Disease (CVD)\\\hspace*{0.5\leftmargin}\labelitemii\hspace{\dimexpr\labelsep+0.5\tabcolsep}Diabetes Mellitus (DM)\\\hspace*{0.5\leftmargin}\labelitemii\hspace{\dimexpr\labelsep+0.5\tabcolsep}Cancer (CA)\\\labelitemi\hspace{\dimexpr\labelsep+0.5\tabcolsep}10 Common Symptoms:\\\hspace*{0.5\leftmargin}\labelitemii\hspace{\dimexpr\labelsep+0.5\tabcolsep}Fever\\\hspace*{0.5\leftmargin}\labelitemii\hspace{\dimexpr\labelsep+0.5\tabcolsep}Edema\\\hspace*{0.5\leftmargin}\labelitemii\hspace{\dimexpr\labelsep+0.5\tabcolsep}Emaciation\\\hspace*{0.5\leftmargin}\labelitemii\hspace{\dimexpr\labelsep+0.5\tabcolsep}Chest Pain\\\hspace*{0.5\leftmargin}\labelitemii\hspace{\dimexpr\labelsep+0.5\tabcolsep}Headache\\\hspace*{0.5\leftmargin}\labelitemii\hspace{\dimexpr\labelsep+0.5\tabcolsep}Abdominal Pain\\\hspace*{0.5\leftmargin}\labelitemii\hspace{\dimexpr\labelsep+0.5\tabcolsep}Hematochezia\\\hspace*{0.5\leftmargin}\labelitemii\hspace{\dimexpr\labelsep+0.5\tabcolsep}Joint Pain\\\hspace*{0.5\leftmargin}\labelitemii\hspace{\dimexpr\labelsep+0.5\tabcolsep}Jaundice\\\hspace*{0.5\leftmargin}\labelitemii\hspace{\dimexpr\labelsep+0.5\tabcolsep}Cough\\
}} & \multirow{12}{=}{Open-ended generation} & \multirow{12}{=}{\makecell[c]{$70$}} \\
& & & \\
& & & \\
& & & \\
& & & \\
& & & \\
& & & \\
& & & \\
& & & \\
& & & \\
& & & \\
& & & \\
\hhline{-|~|--}
\multirow{12}{=}{AI Patient Test Set} & & \multirow{12}{=}{Open-ended interaction} & \multirow{12}{=}{\makecell[c]{$70$}} \\
& & &  \\
& & &\\
& & &  \\
& & & \\
& & & \\
& & & \\
& & & \\ 
& & & \\
& & & \\
& & & \\
& & & \\\hline
\end{tabular}
\end{table}

\subsection{Test Set Construction}
Based on the specially designed evaluation framework, we constructed data sets to reflect the competency of LLMs across various dimensions. An overview of the dataset is presented in Table~\ref{tab:gpbench_overview}. 
All experiments were conducted in Chinese, including the medical records, prompts for LLMs, model responses, scoring criteria, and scoring procedures. This design aligned with the clinical environment and the working language of Chinese medical professionals.
All models were evaluated using default inference parameters and an identical prompt design, which adhered to widely accepted prompt engineering practices. The prompt template incorporated ``role definition, possessed skills, and task objectives.'' The prompts were tuned on a separate development set to ensure the validity of the evaluation on the test sets (see Supplementary Method~\ref{SM1} and Supplementary Method~\ref{SM2}).
The proposed GPBench evaluation dataset consisted of three components: the Multiple-choice Questions (MCQ) Test Set, the Clinical Case Test Set, and the AI Patient Test Set, each of which had a corresponding development set (see Supplementary Method~\ref{SM3}). The following described the construction of the test sets.

The MCQ Test Set comprised 3,661 multiple-choice questions designed to evaluate LLMs’ foundational medical knowledge and theoretical understanding. The Clinical Case Test Set, drawn from 70 real outpatient records, used open-ended case analysis to assess LLMs' ability to develop systematic diagnostic and treatment plans, highlighting their handling of complex clinical problems and potential blind spots. The AI Patient Test Set, also based on the same 70 cases, simulated interactive consultations with AI-generated ``patients'' to evaluate LLMs’ clinical responsiveness and decision-making in realistic outpatient scenarios.

\begin{figure}[htbp]
  \centering
  \includegraphics[width=\textwidth]{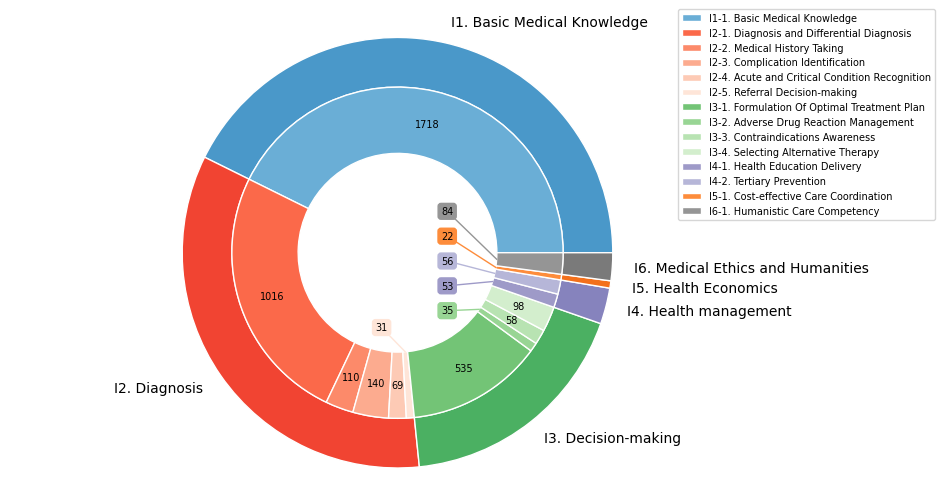}
  \caption{\fontsize{10pt}{10pt}\selectfont The data distribution of Multiple-choice Question (MCQ) Test Set on different competencies.}
\label{fig:data_of_set1}
\end{figure}

The MCQ Test Set was constructed by selecting general practice-related questions from CMB~\cite{wang2023cmb} and MedBench~\cite{MedBench}, and annotating them with secondary indicators from the proposed competency model (see Supplementary Table~\ref{tab:MCQ_example_annotation}). General practice experts manually supplemented them with questions from additional sources to address the problem that some indicators had few associated questions. The final distribution is shown in Fig.~\ref{fig:data_of_set1}.

The Clinical Case Test Set consisted of 70 real-world outpatient records from two medical centers. Based on epidemiological data~\cite{from_jp_1,from_jp_2,from_jp_3}, eight common chronic diseases and ten frequent clinical symptoms were selected, with 3–4 representative cases of varying difficulty levels for each category to ensure comprehensive coverage of typical scenarios (see Supplementary Table~\ref{tab:record_example_annotation} and Supplementary Table~\ref{tab:cad_hard_record}).   
Thus, the cases were distinct from one another, and LLMs were expected to behave very differently.
Strict privacy protection measures were implemented for all medical records used in this study to ensure patient confidentiality. All personally identifiable information—including, but not limited to, names, addresses, phone numbers, national identification numbers, hospital identifiers, physician identifiers, medical record numbers, hospitalization numbers, and specific dates—was systematically removed through a combination of automated scripts and manual review by trained clinical researchers. Specific dates were replaced with time intervals relative to the admission date. The resulting dataset underwent rigorous validation to confirm that no individual could be re-identified.
Expert general practitioners provided detailed annotations in alignment with routine practice, covering differential diagnosis, management, patient education, and recommendations for further examination. General practice specialists subsequently identified key response elements, defined scoring criteria, and mapped each component to the relevant competency dimensions. The outputs from the LLM were evaluated and scored by experienced GPs according to these criteria.

The AI Patient Test Set simulated clinical consultations to efficiently evaluate AI performance. Prompts were designed to elicit natural, conversational responses from patient agents while strictly restricting content to case-provided information (see Supplementary Method~\ref{SM2}). Each ``patient'' only answered questions posed by the ``doctor'' and did not volunteer additional details. A maximum of $T=10$ inquiry rounds were set to assess the model's ability to extract key information within limited interactions.

\subsection{Scoring Protocol}
The clinical cases consisted of anonymized real-world medical records, which were verified and standardized by expert general practitioners (see Supplementary Table~\ref{tab:fever_medium_grountruth_gpt_translation}). Each case was carefully reviewed and refined to enhance its quality, in accordance with standards for medical training. 
Before the evaluation, detailed scoring criteria were developed through a Delphi process to ensure a consistent understanding of all evaluation dimensions and rules. The scoring panel was composed of three licensed GPs, referred to as GP-1, GP-2, and GP-3. GP-1 and GP-2 each had over ten years of clinical experience, while GP-3 was a senior GP with fifteen years of experience. 
All evaluators underwent training on the criteria using a set of randomly selected sample cases to align their interpretations and maintain scoring consistency. During the evaluation, GP-1 and GP-2 independently scored each response. If a discrepancy exceeding 10 points occurred for any individual item, GP-3 conducted arbitration, or the panel convened to reach a consensus (see Supplementary Table~\ref{tab:example_llm_response_results}).

\subsection{Outcome Measures}
\label{outcome_measures}
For the MCQ Test Set, accuracy was calculated based on responses to the corresponding questions. 
Each multiple-choice question was prompted five times for each LLM with different orders of answer options, and the final prediction for each question was determined by majority voting across the five responses.
To conduct a more precise analysis across different competency dimensions, each question was labeled with its corresponding competency indicator. Evaluation was then based on the weighted average accuracy across these indicators.  
The reported weighted average and standard deviation summarize each model’s performance across all questions, reflecting the variance among questions rather than the variance introduced by repeated prompting.

For the Clinical Case Test Set, final scores were assigned by the scoring panel in strict accordance with predefined scoring criteria and the relative weights of each indicator. 
Each case was evaluated once without repeated sampling.
All annotations were independently validated by a separate group of experts. 
The reported weighted average and standard deviation summarize model performance across all cases, reflecting variability in performance among cases.

For the AI Patient Test Set, experts assessed the competency of LLMs in conducting medical history taking (see Supplementary Table~\ref{medical_history_scoring}). Each ``AI patient" agent was evaluated once, and the reported average and standard deviation summarize performance across all agents.

\subsection{Human Evaluation}
To perform a comparative analysis of the performance of LLMs, we conducted a human evaluation benchmark involving general practitioners (see Supplementary Method~\ref{SM4}). Participants were categorized into three groups based on their clinical experience: Group A (more than 10 years of experience, 2 participants), Group B (5 to 10 years of experience, 2 participants), and Group C (less than 5 years of experience, 2 participants).
During the evaluation, 200 multiple-choice questions were selected from the MCQ test set through stratified random sampling to ensure coverage of all competency dimensions. The full set of 70 cases from both the Clinical Case Test Set and the AI Patient Test Set was utilized. All physicians completed the evaluation independently in a quiet environment and received standardized instructions that were nearly identical to those provided to the LLMs.
The scoring process for human responses strictly followed the same procedures and criteria used for evaluating the LLM outputs (see Section~\ref{outcome_measures}), ensuring the results are directly comparable.

\subsection{Statistics \& Reproducibility}
In the MCQ Test Set, the accuracies of the fourteen indicators were assumed to be independent and to follow binomial distributions with varying parameters. Welch's t-test was employed to compare the performance between different LLMs. In the Clinical Case Test Set and AI Patient Test Set, the performance of each LLM was summarized using the mean and standard deviation of the case scores. As the medical records were deliberately selected to include diverse diseases and symptoms,  the score distributions significantly deviated from a Gaussian distribution, as confirmed by the Shapiro-Wilk test ($\alpha= 0.05$). Consequently, the Wilcoxon rank-sum test was used for comparisons between LLMs. All statistical analyses were conducted using Python 3.12.4 and SciPy 1.14.1.

\backmatter

\section*{Data Availability}
The prompts used for LLMs have been included in the manuscript. Both the development set and the test set were made available exclusively for non-commercial research purposes under a data use agreement via our project homepage (\url{https://github.com/AIPrimaryCare}). The development set could be accessed in full, including both the evaluation questions and the corresponding scoring rubrics. To preserve evaluation integrity, only the evaluation questions of the test set were released.
Access requests should be submitted to the corresponding author, accompanied by a detailed description of the research purpose. The corresponding author and the data-contributing institutions will review the request and determine whether access can be granted. Source data are available at the Source Data file.

\section*{Code Availability}
The source code for the study is available in the GitHub repository at \url{https://github.com/AIPrimaryCare/gpbench_code}. To ensure long-term accessibility and reproducibility, the repository has been archived on Zenodo under the DOI 10.5281/zenodo.18428084. A comprehensive README file is provided with full instructions for reproducing the experiments.


\section*{Acknowledgements}
This research was funded by the National Natural Science Foundation of China (72471095), the Guangdong Basic and Applied Basic Research Foundation of China (2024A1515220073), and the Science and Technology Program of Guangzhou (2023B03J1277).

\section*{Author Contribution}
Z. Li wrote the original draft, performed the writing, review, and editing, and developed the methodology. Y.Y. carried out the data curation, performed the formal analysis and visualization, and wrote the original draft, review, and editing. J. Lang performed the writing, review, and editing, carried out the data curation, and conducted the investigation and validation. W.J. conceived the study, developed the methodology, performed the writing, review, and editing, carried out the formal analysis, and supervised the work. J. Chen conducted the investigation and validation, developed the methodology, and performed the writing, review, and editing. Y.Z. carried out the data curation, conducted the investigation, and developed the software. D.W. carried out the data curation, developed the software, and performed the visualization. S. Li conducted the investigation. Z. Lin conducted the investigation. X. Li conducted the investigation. Y.T. conducted the investigation. J.Q. conducted the investigation. X. Lu conducted the investigation. H.Y. conducted the investigation. S. Chen conducted the investigation. Y.B. conducted the investigation. X.Z. conducted the investigation. Y. Chen carried out the data curation. L.Y. conceived the study, developed the methodology, provided the resources, supervised the work, and acquired the funding. Data verification: Y.Y., J. Lang, J. Chen, and Y.Z. directly accessed and verified the underlying data.

\section*{Competing Interests }
All authors declare no competing interests.

\clearpage

\section*{Supplementary Tables}
\setcounter{table}{0}
\renewcommand{\thetable}{T\arabic{table}}
\renewcommand{\tablename}{Supplementary Table}

\fontsize{8pt}{8pt}\selectfont
\begin{longtable}[t!]{>{\raggedright\arraybackslash}m{0.15\linewidth} >{\raggedright\arraybackslash}m{0.17\linewidth} >{\raggedright\arraybackslash}m{0.63\linewidth}}
\caption{The competency indicators and definitions used in our proposed evaluation framework.} \\
    \toprule
    Primary Indicator & Secondary Indicator & Definition \\
    \midrule
    \endfirsthead
    
    \multicolumn{3}{c}{\tablename~\thetable{} -- Continued from previous page} \\
    \midrule
    \textbf{Primary Indicator} & \textbf{Secondary Indicator} & \textbf{Definition} \\
    \midrule
    \endhead
    
    \midrule
    \multicolumn{3}{r}{Continued on next page} \\
    \endfoot
    
    \bottomrule
    \endlastfoot
    \textbf{I1.} Basic Medical Knowledge ($10\%$) & \textbf{I1-1.} Basic Medical Knowledge ($100\%$) & Basic medical knowledge refers to the fundamental scientific understanding that forms the core of medical practice, encompassing the functional, biochemical, and pathological mechanisms of the human body. It serves as the theoretical foundation for clinical reasoning and evidence-based decision-making. \\
    \midrule

    \multirow{5}{*}{ \raisebox{-4.5\baselineskip}[0pt][0pt]{\parbox{\linewidth}{\raggedright\textbf{I2.} Diagnosis ($40\%$)}}} & \textbf{I2-1.} Diagnosis and Differential Diagnosis ($20\%$) & Diagnosis refers to the physician’s judgment of the disease based on the patient’s medical history, symptoms, signs, and auxiliary examination results. 
    Differential diagnosis is the process of distinguishing the patient's main complaint from other diseases and excluding the possibilities of other conditions. 
    Test interpretation belongs to this indicator.
    \\ \cmidrule{2-3}
     & \textbf{I2-2.} Medical History Taking  ($20\%$) & The ability to collect patient information in a comprehensive and accurate manner during the diagnostic and therapeutic process, including the patient's physical symptoms, psychological, mental, social, and cultural factors, as well as family history.
     \\ \cmidrule{2-3}
     & \textbf{I2-3.}  Complication Identification ($20\%$) & The ability to predict, diagnose, and manage potential complications that may arise during the patient's treatment process. Complications refer to other diseases or symptoms that arise during the course of a primary disease, either due to the disease itself or as a result of its treatment. \\ \cmidrule{2-3}
     & \textbf{I2-4.} Acute and Critical Condition Recognition  ($20\%$)& The ability to recognize conditions that occur suddenly, are critical in nature, and require urgent intervention, potentially leading to life-threatening situations. 
     \\ \cmidrule{2-3}
     & \textbf{I2-5.} Referral Decision-making ($20\%$) & The ability to identify and recognize diseases that pose a threat to the patient's life and ensure timely and correct referral to appropriate specialists. \\
    \midrule
    
     \multirow{5}{*}{ \raisebox{-8.5\baselineskip}[0pt][0pt]{\parbox{\linewidth}{\raggedright\textbf{I3.} Decision-making ($30\%$)}}}
     & \textbf{I3-1.} Formulation of Optimal Treatment Plan ($35\%$) & The ability to form an appropriate treatment plan based on the patient's condition. Ancillary test selection is included in this indicator.
     \\ \cmidrule{2-3}
      & \textbf{I3-2.} Adverse Drug Reaction Management ($25\%$) & This indicator refers to the comprehensive capability of healthcare institutions and professionals to promptly identify, scientifically assess, effectively intervene, accurately document, and systematically prevent unintended harmful reactions following medication use in clinical practice. Its core objectives are to minimize drug-related risks, ensure patient safety, and improve clinical outcomes through optimized therapeutic strategies. This competency requires adherence to evidence-based medicine principles, clinical protocols, and regulatory requirements, emphasizing multidisciplinary collaboration and continuous quality improvement. 
      \\ 
     & \textbf{I3-3.} Contraindications Awareness ($25\%$) & The ability to accurately identify, comprehend, and apply drug contraindications (i.e., situations where a specific medication is prohibited or not recommended). This includes knowledge of patient-specific factors (e.g., allergies, pregnancy, hepatic/renal impairment), drug-drug interactions, disease-specific contraindications, and the practical skill to avoid inappropriate medication use in clinical settings, thereby ensuring patient safety.
     \\ \cmidrule{2-3}
     & \textbf{I3-4.}  Selecting Alternative Therapy ($15\%$)  & This indicator refers to the comprehensive ability to systematically evaluate, appropriately choose, and effectively apply non-conventional therapeutic approaches (e.g., acupuncture, herbal medicine, energy therapies) when standard treatments are unsuitable, impractical, or ineffective. This capacity encompasses scientific understanding of alternative therapies, risk-benefit analysis, resource accessibility, and individualized decision-making skills, while integrating cultural context, patient values, and ethical considerations. 
     \\ \midrule
    
    \multirow{5}{*}{ \raisebox{0\baselineskip}[0pt][0pt]{\parbox{\linewidth}{\raggedright\textbf{I4.} Health Management ($10\%$)}}}
    & \textbf{I4-1.}  Health Education Delivery ($50\%$)
 & This competency is required to undertake individual and community health education responsibilities and effectively conduct health education activities.
 \\ \cmidrule{2-3}
     & \textbf{I4-2.} Tertiary Prevention Implementation ($50\%$) & The ability to implement health measures from the perspectives of prevention, treatment, and rehabilitation to achieve ``preventing diseases before they occur, preventing disease progression during illness, and preventing recurrence after illness''. 
     \\ \midrule
    
    \textbf{I5.} Health Economics  ($5\%$) & \textbf{I5-1.} Cost-effective Care Coordination ($100\%$)
& The ability to scientifically control healthcare costs without compromising the quality of medical services, striving to minimize healthcare service fees and resource utilization. \\ \midrule
    
    \textbf{I6.} Medical Ethics and Humanities ($5\%$) & \textbf{I6-1.} Humanistic Care Competency ($100\%$)
 & In medical and healthcare work, the communication between healthcare providers and patients regarding injury, illness, diagnosis, treatment, health, and related factors is primarily led by the healthcare provider. Through various comprehensive means of communication, the goal is to scientifically guide the treatment of the patient’s condition, achieve mutual understanding, and establish a collaborative relationship based on trust, ultimately contributing to human health maintenance, medical development, and societal progress. 
  \label{GPC}
\end{longtable}

\begin{table}[h]
    \centering
    \caption{Parameter settings of evaluated LLMs for GPBench.}
    \fontsize{8pt}{8pt}\selectfont
    \renewcommand{\arraystretch}{1.0}
    \begin{tabular}{>{\centering\arraybackslash}m{0.30\linewidth}>{\centering\arraybackslash}m{0.15\linewidth}>{\centering\arraybackslash}m{0.15\linewidth}>{\centering\arraybackslash}m{0.20\linewidth}}
    \hline
    Model & Temperature & Top-p & Repetition Penalty \\ \hline
    GPT-4-turbo & $1$     & $1$    & $0$  \\
    o1-preview & $1$     & $1$   &  $0$  \\
    Gemini-1.5-pro & Not Reported  & $0.95$  & Not Reported  \\
    Qwen2.5-7B-Instruct & $0.7$    & $0.7$    & $0.5$  \\
    Qwen2.5-72B-Instruct & $0.7$  & $0.7$    & $0.5$ \\
    Claude-3.5-sonnet & $1$    & Not Reported    & Not Reported  \\
    DeepSeek-V3 & $1$  & Not Reported    & Not Reported \\
    DeepSeek-R1 & $1$  & Not Reported    & Not Reported  \\
    HuatuoGPT-o1-7B & $0.3$    & $0.8$   & $1.05$     \\ \hline
    \end{tabular}
\label{tab:detail_parameters}
\end{table}

\begin{table}[h]
    \centering
    \caption{Specializations and information of experts.}
    \begin{tabular}{>{\centering\arraybackslash}p{0.25\linewidth}>{\centering\arraybackslash}p{0.40\linewidth}
                    >{\centering\arraybackslash}p{0.25\linewidth}}
    \hline
    Members & Affiliation & Specialization \\
    \hline
    Jiping Lang (MD)& The Sixth Affiliated Hospital of Sun Yat-sen University & General Practice  \\
    Shuang Li (M.Med)& The Sixth Affiliated Hospital of Sun Yat-sen University & General Practice  \\
    Xuanna Li (M.Med)& The Sixth Affiliated Hospital of Sun Yat-sen University & General Practice  \\
    Yuze Tang (M.Med)& The Sixth Affiliated Hospital of Sun Yat-sen University & General Practice   \\
    Zhu Lin (M.Med)& The Sixth Affiliated Hospital of Sun Yat-sen University & General Practice  \\
    Xiaolin Lu (MBBS)& Xinyi People's Hospital & Cardiology  \\
    Hongji Yu (M.Med)& Xinyi People's Hospital &  Endocrinology    \\
    Jiexian Qiu (MBBS)& Xinyi People's Hospital & Cardiovascular Medicine  \\
    Junrong Chen (M.Med)& The Sixth Affiliated Hospital of Sun Yat-sen University & General Practice \\
    Lin Yao (M.Med) & The Fifth Affiliated Hospital of Sun Yat-sen University & General Practice  \\
    Wenhao Jiang (Ph.D.) & Guangdong Laboratory of Artificial Intelligence and Digital Economy (SZ) & Artificial Intelligence  \\
    Zheqing Li (MSc) & The Sixth Affiliated Hospital of Sun Yat-sen University & Electronic Engineering, Public Health  \\
    \hline
    \end{tabular}
\label{tab:annotators}
\end{table}

\begin{table}[h]
\caption{An example of annotation from the MCQ Test Set.} 
\label{tab:MCQ_example_annotation}
\fontsize{8pt}{8pt}\selectfont
\begin{tabular}{p{0.2\linewidth}p{0.75\linewidth}}
\toprule
Field & Value \\ \midrule
ID             & 1278 \\
Question       & Female, 40 years old, with a 10-year history of rheumatoid arthritis, has been taking generic nonsteroidal anti-inflammatory drugs (NSAIDs) for a long period. Laboratory tests show a hemoglobin level of $78 \ g/L$ in the complete blood count. Which of the following statements regarding this patient's anemia is incorrect? \\
Options        &  \multirow{5}{=}{\makecell[l]{
                A: It is microcytic hypochromic anemia. \\
                B: It belongs to anemia of chronic disease. \\
                C: The primary mechanism is impaired iron utilization. \\
                D: Iron deficiency factors may be involved. \\
                E: It is often accompanied by thrombocytopenia.} }\\
& \\
& \\
& \\
& \\
Answer         & E \\
Type           & single-answer multiple-choice question \\
Source         & cmb-test - Physician Exam - Intermediate-Level Title - Attending Physician of Internal Medicine \\
Indicators     & Diagnosis and Differential Diagnosis (I2-1), Complication Identification (I2-3), Adverse Drug Reaction Management (I3-2)\\ \bottomrule
    \end{tabular}
\end{table}

\begin{table}[h]
\caption{The scoring criteria for the competency in medical history taking on the AI Patient Test Set}
\fontsize{8pt}{8pt}\selectfont
\begin{tabular}{m{0.3\linewidth}|m{0.55\linewidth}|m{0.1\linewidth}}
\hline
\textbf{Primary Dimension} & \textbf{Secondary Dimension} & \textbf{Score}\\ 
\hline
General Information (5 points) & Full and accurate documentation of gender, age, occupation, and other non-identifiable demographic information. & 5  \\ 
\hline
\multirow{7}{=}{History of Present Illness (50 points)} & Onset of disease: time of onset, acuity of progression, potential etiological factors or predisposing causes.  & 6  \\
& Characteristics of the main symptoms or signs: location, nature, duration, severity, alleviating and exacerbating factors. & 10  \\
& Disease progression and clinical course. & 6  \\
& Accompanying symptoms. & 6  \\
& Symptoms relevant for differential diagnosis. & 6  \\
& Diagnosis and treatment process: whether the patient has sought medical attention; examinations performed; medications or treatments prescribed, and their effectiveness. & 10  \\
& General condition and related current medical history. & 6  \\ 
\hline
\multirow{5}{=}{Relevant Medical History (15 points)} & Past medical history. & 3  \\
& Personal history. & 3  \\
& Menstrual history (for pediatrics: birth history). & 3  \\
& Marital and reproductive history (for pediatrics: growth and developmental history). & 3  \\
& Family history. & 3  \\
\hline
\multirow{3}{=}{Clinical Interviewing Skills (20 points)} & Clear structure, with a focus on key points. & 7  \\
& Focus on the patient’s condition, using simple and understandable language. & 7  \\
& Non-suggestive, non-leading, and non-accusatory questioning. & 6  \\
\hline
\multirow{2}{=}{Professionalism (10 points)} & Good service attitude, caring and considerate towards patients. & 5  \\
& Strong communication skills, skillful guidance, no abrupt interruptions of patient narration. & 5  \\
\hline
\end{tabular}
\label{medical_history_scoring}
\end{table}

\clearpage
{\renewcommand{\arraystretch}{1}
\fontsize{8pt}{8pt}\selectfont
\begin{longtable}[l]{p{\textwidth}}
\caption{An example of a fever medical case.}
\label{tab:record_example_annotation} \\

\toprule
\textbf{Case}  \\
\midrule
\endfirsthead

\toprule
\textbf{Case}  \\
\midrule
\endhead

\midrule
\multicolumn{1}{r}{Continued on next page} \\
\endfoot

\bottomrule
\endlastfoot
\textbf{Patient:} male, 47 years old. \\
\textbf{Chief Complaint:} Fever for one week.   \\
\textbf{Present Illness:} One week ago, the patient began to have a fever of $\text{T} \ 38^\circ\text{C}$ after mountain climbing, accompanied by chills and rigors, as well as coughing and expectoration of copious white viscous sputum. He also experienced muscle aches. A complete blood count performed at a local clinic indicated decreased platelets. After treatment, there was no significant improvement in symptoms; the fever recurred with a body temperature rising to $\text{T} \ 39^\circ\text{C}$ along with rigors, dizziness, headache, and occasional nausea. He came to our hospital for further evaluation.   \\
\textbf{Past Medical History:}
He has had stage $2$ hypertension for many years. He is currently not taking any antihypertensive medications, and his blood pressure has remained within normal limits.  \\
\textbf{Physical Examination:} 
$\text{T} \ 38^\circ\text{C}$,$ \text{P} \ 120 $ beats/min, $\text{R} \ 28 $ breaths/min, $\text{BP} \ 136/76 \ mmHg$. 
Conscious and alert. A $0.5 \ cm \times 0.5 \ cm$ skin ulceration was observed on the left lower limb. 
Bilateral lung sounds were slightly diminished, with a small amount of moist rales heard in both lower lungs. 
The heart rate was $120$ beats/min, regular rhythm, and no obvious pathological murmurs were detected in any valve area. The abdomen was flat and soft, with no tenderness or rebound tenderness. 
The liver and spleen were not palpable below the costal margin.
Muscle strength and tone in all four limbs were normal, and no edema was noted in either lower extremity. \\
\textbf{Laboratory and Auxiliary Examinations:} \newline
Complete blood count: $\text{WBC:} \ 8.96 \times 10^{9}/L$, $\text{NEUT:} \ 50.6\%$, $\text{Hb:} \ 150.0 \ g/L$, $\text{PLT:} \ 48 \times 10^{9}/L (\downarrow)$. \newline
$\text{D-dimer:} \ 13.06 \ \mu g/mL (\uparrow)$, $\text{BNP:} \ 69.90 \ pg/mL (\uparrow)$, $\text{hs-cTnI:} \ 0.033 \ ng/mL$, tests for respiratory pathogens, hepatitis B serological markers, and transfusion-transmitted infections were normal. \newline
Biochemistry: normal cardiac enzymes, $\text{BUN:} \ 6.33 \ mmol/L$, $\text{Cr:} 81.50 \mu mol/L$, $\text{Ca:} \ 1.95 \ mmol/L (\downarrow)$, $\text{hs-CRP:} \ 73.52 \ mg/L (\uparrow)$, $\text{ALT:} \ 190.64 \ U/L (\uparrow)$, $\text{AST:} \ 203.98 \ U/L (\uparrow)$, \text{$\gamma$-GT:} $ 173.50 \ U/L (\uparrow)$, $\text{ALB:} \ 28.50 \ g/L (\downarrow)$, $\text{FER:} \ >1025.0 \ ng/mL (\uparrow)$. \newline
$\text{ECG}:$ (1) Sinus rhythm. (2) Low voltage in limb leads. \newline
Cardiac ultrasound: mild mitral regurgitation, mild tricuspid regurgitation, normal left ventricular systolic function. \newline
Abdominal and urinary tract ultrasound: slight splenomegaly; no obvious abnormalities in the liver, gallbladder, bile ducts, pancreas; mildly enlarged prostate; kidneys within normal size range; no ureteral dilation; no abnormal findings in the bladder. \newline
Chest $CT$ report: (1) Multiple scattered inflammatory and fibrotic lesions in both lungs, with a small amount of bilateral pleural effusion. Further follow-up is recommended. (2) Bilateral pleural thickening; possible encapsulated pleural effusion in the right interlobar region, to be correlated clinically.  \\
\textbf{Diagnosis: } \newline
1. Scrub typhus \newline
2. Splenomegaly \newline
3. Liver dysfunction \newline
4. Pleural effusion \newline
5. Hypoproteinemia \newline
6. Secondary thrombocytopenia \newline
7. Stage 2 hypertension, high-risk category  \\
\textbf{Treatment:} \newline
I. General Management \newline
The patient should rest, eat easily digestible foods, maintain oral hygiene, and ensure water and electrolyte balance to reduce complications. For high fever, physical cooling or appropriate use of antipyretic agents may be employed, avoiding excessive sweating caused by certain antipyretics. \newline
II. Etiological Treatment (select one of the following) \newline
1. Tetracyclines: Recommended dose of doxycycline for adults is $0.1 \ g$ twice a day; the first dose is doubled. Continue the anti-infective therapy for at least $3$ days after body temperature returns to normal and clinical symptoms improve and stabilize, generally for $7$–-$10$ days. In severe cases, intravenous administration of doxycycline may be considered: on day $1$, $0.2 \ g$ in $1$–-$2$ infusions, then $0.1$–-$0.2$ $g$ daily depending on infection severity. \newline
2. Macrolides: Roxithromycin $150 \ mg$ twice a day for adults; after fever subsides, $150 \ mg$ once a day. Azithromycin $0.5 \ g$ once daily for adults; after fever subsides, $0.25 \ g$ once daily. Clarithromycin $0.5 \ g$ once every $12$ hours for adults. All of these regimens are given for $7$–-$10$ days. \newline
III. Symptomatic and Supportive Treatment \newline
In cases of scrub typhus complicated by respiratory, hepatic, or renal dysfunction, treatments may include anti-inflammatory therapy, oxygen supplementation, liver protection and enzyme-lowering measures, as well as maintenance or replacement therapy for renal function.  \\
\end{longtable}
}

\clearpage
\begin{table}[t]
\caption{An example of annotated scoring criteria for the sample of a fever medical record.}
\label{tab:fever_medium_grountruth_gpt_translation}
{\fontsize{8pt}{8pt}\selectfont
\begin{tabular}{p{0.2\linewidth}p{0.2\linewidth}p{0.55\linewidth}}
\toprule
\textbf{Indicator} & \textbf{Applicability} & \textbf{Scoring Criteria} \\
\midrule
Diagnosis and Differential Diagnosis & Yes &
1. Scrub typhus (40 points)\newline
2. Splenomegaly (10 points)\newline
3. Hepatic insufficiency (10 points)\newline
4. Pleural effusion (10 points)\newline
5. Hypoproteinemia (10 points)\newline
6. Secondary thrombocytopenia (10 points)\newline
7. Grade 2 hypertension; high-risk group (10 points) \\[1ex]

Referral Decision-making & Yes &  Referral to the department of infectious diseases (100 points). \\[1ex]

Acute and Critical Condition Recognition & Yes &  Hepatic insufficiency (25 points); Pleural effusion (25 points); Hypoproteinemia (25 points); Secondary thrombocytopenia (25 points) \\[1ex]

Complication Identification & Yes & No critical or severe conditions were involved (100 points). \\[1ex]

Formulation of Optimal Treatment Plan & Yes & 
\textbf{Part 1. Optimal treatment plan} \newline 
I. Etiological Treatment (choose one medication if appropriate) (50 points; if the medication name or usage and dosage are incorrect, deduct 30 points)\newline
1. Tetracyclines: Doxycycline \newline
   Recommended dosage for adults: 0.1 g twice a day, with the initial dose doubled. The anti-infective course should continue for at least 3 days after body temperature returns to normal and clinical symptoms have improved and stabilized, generally for 7–10 days. For severe cases, intravenous infusion may be selected. On the first day, adults receive 0.2 g in 1–2 infusions. Thereafter, according to the severity, 0.1–0.2 g is administered intravenously per day.\newline
2. Macrolides: Roxithromycin 150 mg twice a day for adults; once the fever subsides, 150 mg once a day. Azithromycin 0.5 g once a day for adults; after defervescence, 0.25 g once a day. Clarithromycin 0.5 g per dose for adults, administered orally once every 12 hours. All of these medications are given for a total of 7–10 days.\newline
II. Symptomatic and Supportive Treatment (50 points)\newline
When scrub typhus is complicated by respiratory, hepatic, or renal dysfunction, measures often include anti-inflammatory therapy, oxygen supplementation, liver-protective interventions with enzyme-lowering therapy, and maintenance or replacement of renal function.\newline 
\textbf{Part 2. Clinical appropriateness evaluation of diagnostic examinations} \newline
N/A \\[1ex]

Health Education Delivery & Yes & 
Patients should ensure adequate rest, consume easily digestible foods, pay attention to oral hygiene, maintain fluid and electrolyte balance, and reduce the risk of complications. Physical cooling measures may be used for high fever, and antipyretic medications can be administered as appropriate. However, caution should be exercised in the use of fever-reducing drugs that induce excessive sweating(100 points).
 \\[1ex]
Cost-effective Care Coordination & No & N/A \\[1ex]
Humanistic Care Competency & No & N/A \\
\bottomrule
\end{tabular}}
\end{table}

\clearpage
{\renewcommand{\arraystretch}{1}
\fontsize{8pt}{8pt}\selectfont
\begin{longtable}[l]{p{0.95\linewidth}}
\caption{An example of a \textit{CAD} medical case.}
\label{tab:cad_hard_record} \\

\toprule
\textbf{Case}  \\
\midrule
\endfirsthead

\toprule
\textbf{Case}  \\
\midrule
\endhead

\midrule
\multicolumn{1}{r}{Continued on next page} \\
\endfoot

\bottomrule
\endlastfoot
\textbf{Patient:} male, $74$ years old. \\
\textbf{Chief Complaint:} Intermittent chest pain for $2$ years, worsened over the past week. \\
\textbf{Present Illness:} Two years ago, the patient began experiencing dull chest pain while walking briskly. The pain was mild, located behind the lower sternum, about the size of a fist, and lasted for approximately $5$ minutes, gradually improving with rest. There was no radiation to the left arm or shoulder/neck region, and no relation to food intake or emotional triggers. No nighttime occurrences were reported. At that time, there were no associated symptoms like dizziness, blackouts, syncope, fatigue, sweating, dyspnea, oliguria, edema, fever, cough, hemoptysis, nausea, vomiting, abdominal pain, or skin abnormalities. The patient had no trauma, prolonged bed rest, fractures, or surgical history. An ECG at a local clinic was normal, and blood pressure was within normal range. No further tests were done, and symptoms did not recur until one week ago. \newline
During exercise one week ago, the chest pain recurred with increased severity, lasting about $10$ minutes and improving with rest. It was accompanied by sweating but no dizziness or blackout. Blood pressure and ECG remained normal. Five days ago, the patient visited the Sixth People’s Hospital, where coronary CTA showed approximately $70\%$ stenosis in the proximal left anterior descending artery. Today, he came to our clinic seeking further diagnostic and treatment advice. He is alert, in good spirits, has a normal diet and sleep, normal bowel and urinary function, and no weight change. \\
\textbf{Past Medical History:} \newline
The patient was diagnosed with hypertension $15$ years ago, with a maximum recorded BP of $170/100 \ mmHg$ ($1 \ mmHg = 0.133 \ kPa$). He takes irbesartan/hydrochlorothiazide $150/12.5 \ mg$ once daily, with blood pressure maintained around $120$–$130/70$–$80 \ mmHg$. No history of cerebrovascular disease, diabetes, or dyslipidemia. Vaccination history unknown. No surgical or trauma history, no drug or food allergies, no history of blood transfusion, hepatitis, or contact with endemic areas. \\
\textbf{Family History:} \newline
No known hereditary diseases in the family. \\
\textbf{Personal History:} \newline
Non-smoker, non-drinker. Daily salt intake $8$–$10 \ g$, staple food around $300 \ g$, oil $40 \ g$, meat/egg $200 \ g$. Engages in brisk walking or jogging for approximately $45 \ min$ daily. Harmonious family life, good financial situation. Junior college education, lives with spouse after retirement. Married, with one daughter. \\
\textbf{Physical Examination:} \newline
$\text{T} \ 36.9^\circ\text{C}$, $\text{P} \ 88$ beats/min, $\text{R} \ 18$ breaths/min, $\text{BP} \ 137/80 \ mmHg$, height $158 \ cm$, weight $52 \ kg$, waist circumference $86 \ cm$, BMI $21 \ kg/m^2$. Conscious, symmetrical nasolabial folds, midline tongue protrusion, soft neck, midline trachea. Lungs resonant to percussion, clear breath sounds, no rales or pleural rubs. Cardiac apex at the fifth intercostal space, $0.5 \ cm$ inside left midclavicular line, size approx. $2 \ cm$, no thrills or pericardial rubs. Heart boundaries normal, heart rate $88$ beats/min, regular rhythm, no murmurs or extra sounds. Abdomen soft, non-tender, no rebound pain, no vascular murmurs heard. No lower limb edema. Dorsalis pedis pulses palpable. Muscle strength and tone in all limbs normal. \\
\textbf{Laboratory and Auxiliary Examinations:} \newline
ECG: normal. \newline
Chest X-ray: no significant abnormalities. \newline
Coronary CTA: approximately $70\%$ stenosis in the proximal left anterior descending artery. \\
\textbf{Diagnosis:} \newline
1. Atherosclerotic coronary artery disease (effort angina, CCS class II, high risk) \newline
2. Hypertension stage $2$ (very high-risk group) \\
\textbf{Treatment:} \newline
I. Further Examinations \newline
Complete blood count, coagulation profile, blood glucose, lipid panel, liver and renal function, BNP, homocysteine, urinalysis; $24h$ ambulatory blood pressure monitoring, 24h Holter ECG, echocardiography, vascular ultrasound (carotid and lower extremities), fundoscopy for end-organ damage and atherosclerosis assessment. \newline
II. Non-pharmacological Treatment \newline
1. Diet: low salt ($<6$g/day), low fat ($<25 $g/day), light and easily digestible foods rich in vitamins. Avoid fried food, animal organs, and cholesterol-rich food. Prefer small, frequent meals; avoid overeating. No spicy food, alcohol, tobacco, strong tea, or coffee. \newline
2. Physical activity: mild walking before coronary angiography; after procedure, gradually resume moderate aerobic exercise ($30$–$45 \ min$, $5$ times/week) depending on condition. \newline
3. Self-monitoring of BP and heart rate daily, record symptoms (chest pain, tightness, foamy urine), take medications regularly. Target BP $<140/90 \ mmHg$, ideally $<130/80 \ mmHg$ if tolerated (avoid excessive diastolic drop). \newline
4. Psychological support: help reduce stress, ensure adequate sleep, encourage adherence to treatment plan. \newline
5. Regular follow-up: every $1$–$3$ months to monitor drug side effects. \newline
6. Risk assessment via coronary angiography: for patients with mild angina (CCS I–II) or asymptomatic, yet high-risk by non-invasive stratification, consider revascularization. Recommend this patient undergo coronary angiography in a higher-level hospital; stenting if necessary. \newline
III. Pharmacological Treatment \newline
1. Heart rate control: e.g., metoprolol tartrate $50 \ mg$ twice daily, target HR $55$–$60$/min. \newline
2. Coronary vasodilation: e.g., isosorbide mononitrate capsule $40 \ mg$ once daily. \newline
3. Lipid regulation: e.g., atorvastatin $20 \ mg$ once every evening, LDL $<1.8 \ mmol/L$. \newline
4. Antiplatelet therapy: aspirin $100 \ mg$ once every evening. \newline
5. Antihypertensive: e.g., irbesartan/hydrochlorothiazide $150/12.5 \ mg$ once daily; maintain BP $<130/80 \ mmHg$ if tolerated. \newline
6. Follow-up regularly; seek care if symptoms occur. \\

\label{llm_generates_case}
\end{longtable}
}

\clearpage
{\renewcommand{\arraystretch}{1}
\fontsize{8pt}{8pt}\selectfont
\begin{longtable}[l]{p{0.95\linewidth}}
\caption{o1-preview's response to the above \textit{CAD} medical case and its evaluation results.}
\label{tab:example_llm_response_results} \\

\toprule
\textbf{Detailed Evaluation Results} \\
\midrule
\endfirsthead

\toprule
\textbf{Detailed Evaluation Results}  \\
\midrule
\endhead

\midrule
\multicolumn{1}{r}{Continued on next page} \\
\endfoot

\bottomrule
\endlastfoot
\textbf{Indicator:} Diagnosis and Differential Diagnosis\\\textbf{Applicability:} Yes\\\textbf{Scoring Criteria:}\\1. Coronary Atherosclerotic Heart Disease ($30$ points): Effort-induced angina, CCS Class II, high risk ($40$ points)\\2. Grade $2$ Hypertension (Very high-risk group) ($30$ points)\\\textbf{LLM Response:}\\The patient's clinical manifestations and auxiliary examination results suggest the possible presence of stable angina, primarily caused by coronary artery stenosis. Coronary CTA shows approximately $70\%$ stenosis in the proximal segment of the left anterior descending artery, which is consistent with the presentation of coronary artery disease, particularly the diagnosis of stable angina.\\\textbf{Score:} $60$\\
\midrule
\textbf{Indicator:} Complication Identification\\\textbf{Applicability:} Yes\\\textbf{Scoring Criteria:}\\No complications are involved.~($100$ points)\\\textbf{LLM Reponse:}\\No obvious complications are currently observed. However, given the patient's history of hypertension, attention should be paid to its long-term effects on the cardiovascular system, such as the potential risk of heart failure or other cardiovascular events.\\\textbf{Score:} $100$\\
\midrule
\textbf{Indicator:} Acute and Critical Condition Recognition\\\textbf{Applicability:} Yes\\\textbf{Scoring Criteria:}\\No acute or critical conditions are involved. ($100$ points)\\\textbf{LLM Response:}\\The current condition does not qualify as an acute or critical illness. The patient’s symptoms gradually improve with rest, and there are no severe manifestations such as worsening chest pain, dyspnea, or syncope. However, if such symptoms occur, further emergency evaluation should be considered.\\\textbf{Score:} $100$\\
\midrule
\textbf{Indicator:} Referral Decision-making\\
\textbf{Applicability:} Yes\\
\textbf{Scoring Criteria:}\\
It is recommended that the patient be referred to a higher-level hospital for coronary angiography. If necessary, stent implantation should be performed. ($100$ points)\\
\textbf{LLM Response:}\\Referral to the cardiology department is recommended for further evaluation and treatment. Additional tests such as echocardiography, stress testing, or further coronary angiography may be necessary to determine whether interventional therapy (e.g., stent implantation) or adjustments to medication are required.\\
\textbf{Score:} $100$\\
\midrule
\textbf{Indicator:} Formulation of Optimal Treatment Plan\\
\textbf{Applicability:} Yes\\
\textbf{Part 1. Optimal treatment plan}\\
\textbf{Scoring Criteria:}\\
1. Pharmacological Treatment (Half of the corresponding points will be deducted for incorrect usage or dosage): Heart rate control (Target heart rate: $55$–$60$ bpm, e.g., Metoprolol tartrate $50 mg$~orally twice daily or Metoprolol succinate $23.75 mg$~orally once daily) ($15$ points); Coronary vasodilation (e.g., Isosorbide mononitrate capsules $40 mg$ orally once daily) ($15$ points); Lipid-lowering therapy (LDL-C target $<1.8 mmol/L$, e.g., Atorvastatin $20 mg$ orally once every evening or Rosuvastatin calcium $10 mg$ orally once every evening) ($20$ points); Antiplatelet therapy (e.g., Aspirin $0.1 g$ orally once every evening or Clopidogrel hydrogen sulfate $100 mg$ orally once daily) ($20$ points); Antihypertensive therapy (If tolerated, blood pressure target $<130/80 mmHg$, e.g., Irbesartan/hydrochlorothiazide $150 mg/12.5 mg$ orally once daily or Irbesartan/amlodipine $100 mg/5 mg$ orally once daily) ($20$ points)\\
2. Regular Follow-up: Follow-up every $1–3$ months with monitoring of drug side effects ($10$ points)\\
\textbf{LLM Response:}\\
Optimize pharmacological treatment by using aspirin $100 mg$ once daily to prevent thrombosis, atorvastatin $20 mg$ once daily to control blood lipids, metoprolol $50 mg$ twice daily to reduce cardiac workload, and nitroglycerin sustained-release tablets $30 mg$ once daily to relieve angina. Additionally, coronary interventional therapy, such as stent implantation, is recommended to improve blood flow.\\
\textbf{Score:}$70$\\
\textbf{Part2. Clinical appropriateness evaluation of diagnostic examinations}\\
\textbf{Scoring Criteria:}\\
Complete blood count ($5$ points), coagulation profile ($5$ points), blood glucose ($10$ points), blood lipids ($10$ points), liver and kidney function ($10$ points), BNP ($5$ points), homocysteine ($5$ points), urinalysis ($5$ points); $24$-hour ambulatory blood pressure monitoring ($10$ points), $24$-hour Holter monitoring ($10$ points), echocardiography ($10$ points), ultrasound of neck and lower limb vessels ($10$ points), and fundus examination ($5$ points) to evaluate for atherosclerosis in other areas and target organ damage.\\
\textbf{LLM Response:}\\
It is recommended to perform coronary angiography to thoroughly assess the degree and extent of stenosis, conduct echocardiography to evaluate cardiac function, and carry out laboratory tests such as blood lipids and blood glucose to comprehensively assess cardiovascular risk.\\
\textbf{Score:} $30$\\
\midrule
\textbf{Indicator:} Health Education Delivery \\
\textbf{Applicability:} Yes\\
\textbf{Scoring Criteria:}\\
1. Healthy diet: A low-salt, low-fat diet is recommended ($20$ points), with daily salt intake $<6g$ and cooking oil $<25g$. The diet should be light, easy to digest, and rich in vitamins. Avoid fried foods, animal organs, and other high-cholesterol foods. Eat small, frequent meals and avoid overeating. Spicy foods should be avoided, and smoking, alcohol, strong tea, and coffee should be prohibited to prevent excessive stimulation and increased heart rate. ($20$ points)\\
2. Appropriate physical activity: Before coronary angiography, light walking is recommended while avoiding strenuous exercise. After the procedure, moderate-intensity aerobic exercise ($30–45$ minutes per session, $5$ times/week) should be resumed gradually based on the patient’s condition. ($20$ points)\\
3. Daily monitoring: Measure and record blood pressure and heart rate daily. Be alert to symptoms such as chest pain, chest tightness, and foamy urine. Seek timely medical consultation and adhere strictly to medication schedules. For patients with coronary artery disease and hypertension, a blood pressure target of $<140/90 mmHg$ is recommended; if tolerated, aim for $<130/80 mmHg$ while avoiding excessively low diastolic pressure. ($20$ points)\\
4. Psychological support: Help patients prevent and relieve mental stress, maintain sufficient sleep, and actively cooperate with the prescribed treatment plan. ($20$ points)\\
\textbf{LLM Response:}\\
Instruct the patient to follow a low-salt, low-fat diet; encourage regular and moderate physical activity; educate the patient to take medications on time and regularly monitor blood pressure and heart rate; teach the patient to recognize symptoms of angina and seek medical attention promptly.\\
\textbf{Score:} $60$\\
\midrule
\textbf{Indicator:} Cost-effective Care Coordination\\
\textbf{Applicability:} No\\
\textbf{Score:} N/A\\
\midrule
\textbf{Indicator:} Humanistic Care Competency\\
\textbf{Applicability:} No\\
\textbf{Score:} N/A
\end{longtable}
}

\begin{sidewaystable}
\caption{The accuracies (\%) of LLMs on the MCQ Test set across the secondary competency indicators.}\label{results_test_1}
\begin{tabular}{m{0.09\textheight}m{0.17\textheight}m{0.05\textheight}m{0.05\textheight}m{0.04\textheight}m{0.05\textheight}m{0.05\textheight}m{0.05\textheight}m{0.05\textheight}m{0.07\textheight}m{0.05\textheight}m{0.05\textheight}}
\toprule
\textbf{Primary Indicator} & \textbf{Secondary Indicator} & Qwen2.5-7B-Instruct & Qwen2.5-72B-Instruct & GPT-4o & GPT-4-turbo & o1-preview & Claude-3.5-Sonnet & Gemini-1.5-pro & HuatuoGPT-o1-7B & DeepSeek-V3 & DeepSeek-R1 \\
\midrule
I1. Basic Medical Knowledge ($10\%$)  & I1-1.  Basic Medical Knowledge ($100\%$)  & $64.48$ & $76.60$ & $68.83$ & $54.53$ & $80.01$ & $53.73$ & $64.98$ & $70.59$ & $71.86$ & $82.44$\\ 
\midrule
\multirow{5}{=}[-36pt]{\makecell[c]{I2. Diagnosis \\ ($40\%$)}}  & I2-1. Diagnosis and Differential Diagnosis ($20\%$)  & $68.90$ & $79.80$ & $74.36$ & $60.86$ & $82.18$ & $59.36$ & $68.80$ & $74.28$ & $77.00$ & $85.38$\\
\cmidrule{2-12}
 & I2-2. Medical History Taking ($20\%$)  & $71.82$ & $79.82$ & $77.45$ & $59.27$ & $85.45$ & $61.09$ & $68.55$ & $74.18$ & $80.00$ & $85.82$\\
 \cmidrule{2-12}
 &I2-3. Complication Identification ($20\%$)  & $65.95$ & $80.83$ & $73.39$ & $62.66$ & $83.12$ & $55.65$ & $69.38$ & $73.24$ & $78.97$ & $89.27$\\
 \cmidrule{2-12}
 &I2-4. Acute and Critical Condition Recognition ($20\%$)  & $77.68$ & $82.90$ & $82.32$ & $63.77$ & $86.38$ & $59.13$ & $69.28$ & $82.03$ & $82.03$ & $87.54$\\
 \cmidrule{2-12}
 &I2-5. Referral Decision-making ($20\%$)  & $65.82$ & $70.89$ & $64.56$ & $72.15$ & $62.03$ & $72.78$ & $73.42$ & $60.51$ & $68.35$ & $61.39$\\ 
 \midrule
\multirow{4}{=}[-18pt]{\makecell[c]{I3. Decision\\-making ($30\%$)}}  &I3-1. Formulation of Optimal Treatment Plan ($35\%$) & $68.35$ & $76.88$ & $65.35$ & $51.75$ & $74.62$ & $50.65$ & $61.10$ & $72.12$ & $73.88$ & $82.58$\\
\cmidrule{2-12}
 &I3-2.  Adverse Drug Reaction Management ($25\%$)  & $66.86$ & $77.71$ & $69.71$ & $58.86$ & $75.43$ & $48.57$ & $63.43$ & $75.43$ & $74.86$ & $80.57$\\
 \cmidrule{2-12}
 &I3-3.  Contraindications Awareness ($25\%$)  & $59.31$ & $74.83$ & $69.66$ & $46.55$ & $81.38$ & $48.97$ & $64.48$ & $67.59$ & $76.55$ & $82.76$\\
 \cmidrule{2-12}
 &I3-4.  Selecting Alternative Therapy ($15\%$) & $64.12$ & $74.12$ & $63.53$ & $54.71$ & $72.35$ & $50.00$ & $60.00$ & $66.67$ & $71.18$ & $79.12$\\
 \midrule
\multirow{2}{=}{\makecell[c]{I4. Health \\ Management \\($10\%$)}}  & I4-1.  Health Education Delivery ($50\%$) & $74.34$ & $78.11$ & $77.36$ & $72.08$ & $86.04$ & $68.68$ & $76.23$ & $76.89$ & $76.60$ & $84.53$\\
\cmidrule{2-12}
 & I4-2.  Tertiary Prevention Implementation ($50\%$) & $79.29$ & $81.07$ & $72.86$ & $67.14$ & $86.07$ & $57.50$ & $73.21$ & $83.57$ & $76.07$ & $89.64$\\
 \midrule
I5. Health Economics ($5\%$)  & I5-1. Cost-effective Care Coordination ($100\%$) & $79.00$ & $84.00$ & $76.00$ & $74.00$ & $80.00$ & $71.00$ & $73.00$ & $81.82$ & $80.00$ & $89.00$\\
\midrule
I6. Medical Ethics and Humanities ($5\%$)  & I6-1. Humanistic Care Competency ($100\%$) & $80.14$ & $78.49$ & $73.29$ & $67.61$ & $75.41$ & $67.38$ & $78.72$ & $86.76$ & $75.18$ & $82.03$\\
\midrule
\multicolumn{2}{c}{Weighted Average} & $69.63$ & $78.13$ & $71.80$ & $60.79$ & $79.16$ & $58.13$ & $68.22$ & $73.95$ & $75.80$ & $82.74$\\
\multicolumn{2}{c}{STD} & $1.40$ & $1.28$ & $1.38$ & $1.45$ & $1.27$ & $1.47$ & $1.41$ & $1.34$ & $1.32$ & $1.20$ \\
\botrule
\end{tabular}
\end{sidewaystable}

\newpage

\begin{table}[htbp]
\caption{The performance (scores) of LLMs on the Clinical Case Test Set.}\label{results_cases}
\fontsize{8pt}{8pt}\selectfont
\begin{tabular}{m{0.1\textwidth}|m{0.1\textwidth}m{0.1\textwidth}m{0.1\textwidth}m{0.1\textwidth}m{0.1\textwidth}m{0.1\textwidth}|m{0.08\textwidth}m{0.05\textwidth}}
\hline
Model & Diagnosis and Differential Diagnosis (I2-1) & Complication Identification (I2-3) & Acute and Critical Condition Recognition (I2-4) & Referral Decision-making (I2-5) & Formulation of Optimal Treatment Plan (I3-1) & Health Education Delivery (I4-1) & Weighted Average & STD\\
\hline 
Qwen2.5-7B-Instruct & $63.36$ & $62.00$ & $80.86$ & $75.71$ & $56.60$ & $82.07$ & $68.63$ & $16.26$\\
Qwen2.5-72B-Instruct & $67.57$ & $60.36$ & $88.14$ & $74.14$ & $66.43$ & $87.71$ & $72.79$ & $17.18$\\
GPT-4o & $70.71$ & $73.79$ & $87.14$ & $81.43$ & $64.81$ & $85.21$ & $76.02$ & $19.29$\\
GPT-4-turbo & $62.13$ & $67.29$ & $83.71$ & $78.86$ & $60.11$ & $81.71$ & $71.07$ & $20.13$\\
Claude-3.5-Sonnet & $69.21$ & $73.00$ & $85.00$ & $77.57$ & $62.06$ & $79.50$ & $73.42$ & $18.33$\\
Gemini-1.5-pro & $76.36$ & $72.57$ & $85.00$ & $78.86$ & $65.93$ & $87.43$ & $76.46$ & $20.14$\\
DeepSeek-V3 & $63.14$ & $ 69.29$ & $83.57$ & $74.00$ & $59.79$ & $84.00$ & $70.90$ & $21.93$\\
DeepSeek-R1 & $78.93$ & $76.57$ & $94.29$ & $79.84$ & $77.04$ & $87.93$ & $81.80$ & $16.92$\\
HuatuoGPT-o1-7B & $52.79$ & $70.29$ & $87.86$ & $72.14$ & $60.00$ & $89.43$ & $70.35$ & $19.88$\\
o1-preview & $67.64$ & $68.86$ & $85.71$ & $85.29$ & $61.09$ & $85.00$ & $74.24$ & $20.26$ \\
\hline
\end{tabular}
\end{table}

\begin{table}[h]
    \centering
    \caption{The results of human physicians and LLMs on the evaluation dataset.}
    \label{tab:human_123}
    {
    \fontsize{8pt}{10pt}\selectfont
    \begin{tabular}{
    >{\arraybackslash}m{0.35\linewidth}
    >{\arraybackslash}m{0.18\linewidth}
    >{\arraybackslash}m{0.18\linewidth}
    >{\arraybackslash}m{0.18\linewidth}}
    \hline
    Human Physicians / Models & MCQ Test Set (accuracy $\%$) & Clinical Case Test Set (score) & AI Patient Test Set (score) \\ \hline
    Group A & 62.97 & 87.10 & 77.75 \\
    Group B  &62.39& 85.77& 76.59\\
    Group C &68.32& 81.16& 61.30 \\ \hline
    GPT-4o~\cite{hurst2024gpt4o} &71.80 &76.02 &51.09  \\
GPT-4-turbo &60.79 &71.07 &42.06  \\
o1-preview~\cite{openaio1}  & 79.16 &74.24&55.20 \\
Gemini-1.5-pro~\cite{gemini1.5} &68.22 &76.46 &47.10\\
Qwen2.5-7B-Instruct~\cite{qwen2.5}&69.63 &68.63 & 37.13\\
Qwen2.5-72B-Instruct~\cite{qwen2.5}&78.13 &72.79 &44.01  \\
Claude-3.5-Sonnet  &58.13 &73.42 &53.67\\
DeepSeek-V3~\cite{deepseekv3} &75.80 &70.90 &37.14 \\
DeepSeek-R1~\cite{deepseekr1} &82.74&81.80 & - \\
HuatuoGPT-o1-7B~\cite{huatuogpto1}  &73.95 &70.35 & 36.33\\ 
\hline
    \end{tabular}
    }
\end{table}

\clearpage

\section*{Supplementary Methods}

\supmethod{Prompt for evaluation on the Clinical Case Test Set}{SM1}
\begin{tcolorbox}[colback=blue!5!white,
colframe=black,
width=1.0\textwidth,
title={Diagnostic Plan Prompt},]
\fontsize{8pt}{8pt}\selectfont
\textbf{Role:} You are a general practitioner.\\  
\textbf{Skills:} Possess professional knowledge, excellent communication skills, and attention to detail. Capable of identifying referral needs, recognizing critical and severe conditions, identifying complications, and making accurate diagnoses.\\  
\textbf{Task:} Based on the medical record, physical examination results and auxiliary examination results, provide a diagnosis for the patient.\\  
\textbf{Patient Medical Record:} [\textbf{Patient Medical Record}]\\  
\textbf{Physical Examination Results:} [\textbf{Physical Examination Results}]\\
\textbf{Auxiliary Examination Results:} [\textbf{Auxiliary Examination Results}]\\  
\\
\textbf{Requirements:}\\  
- The doctor needs to determine whether the patient requires a referral and specify the department for referral; determine whether the condition is critical or severe; identify any complications; and provide a detailed and specific diagnosis.\\  
- The output must follow the specified format. The diagnostic content should be described in words.\\  
- Different pieces of information in the examination results should be separated by semicolons.\\  
- Based on the symptoms and medical history provided by the patient, diagnose the condition in detail and with specificity.\\  
- The diagnosis should specify the type and severity level of the condition.\\  
- If a specific diagnosis cannot be made based on the available information, provide suspected conditions.\\  
- The severity and stage of the condition should be represented using Arabic numerals.\\  
- Output only in JSON format, no additional content.\\  
\end{tcolorbox}

\begin{tcolorbox}[colback=blue!5!white,
colframe=black,
width=1.0\textwidth,
title={Treatment Plan Prompt},]
\fontsize{8pt}{8pt}\selectfont
\textbf{Role:} You are a general practitioner.\\  
\textbf{Skills:} Possess professional knowledge, excellent communication skills, and attention to detail. Capable of providing optimal treatment plans, recommending further examinations, offering health education, and ensuring cost-effective treatment.\\  
\textbf{Task:} Based on the medical record, auxiliary examination results, physical examination results, and diagnosis, provide a decision-making plan for the patient.\\  
\textbf{Patient Medical Record:} [\textbf{Patient Medical Record}]\\  
\textbf{Auxiliary Examination Results:} [\textbf{Auxiliary Examination Results}]\\  
\textbf{Physical Examination Results:} [\textbf{Physical Examination Results}]\\  
\textbf{Diagnosis:} [\textbf{Diagnosis}]\\
\\
\textbf{Requirements:}\\  
- The output must follow the specified format.\\  
- The decision-making plan should be divided into three parts: optimal treatment plan, further examinations, and health education.\\  
- The content of the optimal treatment plan, further examinations, health education, and cost-effective treatment should be described in words.\\  
- Different pieces of information in the optimal treatment plan, further examinations, health education, and cost-effective treatment should be separated by semicolons.\\  
- The optimal treatment plan should include specific drug names, dosages, and the purpose of drug therapy.\\  
- Output only in JSON format, no additional content.\\
\end{tcolorbox}

\clearpage

\supmethod{Prompt for evaluation on the AI Patient Test Set}{SM2}
\begin{tcolorbox}[
    colback=blue!5!white,
    colframe=black,
    width=1.0\textwidth,
    title={Patient Agent prompt},
    fonttitle=\bfseries,
]
\fontsize{8pt}{8pt}\selectfont

\textbf{Role:} You are a patient visiting a hospital.\\
\textbf{Patient Information:} [\textbf{Patient Information}]\\
\textbf{Task}: Respond to doctor's questions strictly according to the following categories and action guidelines to simulate realistic doctor-patient dialogue.\\
\textbf{Goal}: Make the conversation as close as possible to real patient-doctor interactions.\\

\textbf{Question Types and Action Guidelines:}\\
-\textbf{Initialization:} Doctor's first question. Only briefly answer about your most troubling symptoms based on the "Chief Complaint" in your personal information.\\
-\textbf{Conclusion:} When the doctor makes a diagnosis or reaches maximum dialogue length, remain silent and wait for further instructions.\\
-\textbf{Valid Inquiries:} If the doctor asks for information in your profile (including "Chief -Complaint", "Present Illness History", "Past Medical History", etc.), provide the corresponding information directly and concisely.\\
-\textbf{Invalid Inquiries:} For any questions not covered in your profile, always answer "I'm not sure."\\
-\textbf{Valid Suggestions:} If the doctor recommends tests or examinations with results included in your profile, confirm and respond with those results.\\
-\textbf{Invalid Suggestions:} For suggested tests or examinations not in your profile, express willingness to follow these recommendations.\\

\textbf{Special Notes:}\\
-Keep responses brief and clear, do not repeat doctor's suggestions or conclusions. \\
-Use natural, conversational language to make dialogue more authentic.\\
-Do not mention any personal information unless specifically asked. \\
-Answers should directly address the doctor's questions, avoid providing irrelevant information.\\
-Do not mention symptoms or conditions not listed in your personal information. \\
-Do not mention anything not included in your personal information.\\
-When uncertain about an answer, consistently respond with ``I'm not sure" \\
-Do not fabricate any false examination results.\\

Once the consultation begins, please respond to each of the doctor's questions one by one.\\
\end{tcolorbox}

\begin{tcolorbox}[
    colback=blue!5!white,
    colframe=black,width=1.0\textwidth,
    title={Doctor Agent prompt},
    fonttitle=\bfseries,
]
\fontsize{8pt}{8pt}\selectfont

\textbf{Role:} You are a general practitioner.  \\
\textbf{Skills:} General practitioners possess the ability to diagnose and differentiate diagnoses, collect medical histories, identify complications, recognize critical and severe conditions, determine the need for referrals, provide optimal treatment plans, manage adverse drug reactions, suggest alternative treatments, educate patients on health topics, implement three levels of prevention, enhance patient compliance, offer family support, control treatment costs reasonably, and provide humanistic care.  \\
\textbf{Task:} Conduct a medical consultation.  \\
\textbf{Objective:} Gather as much information as possible regarding the patient's chief complaint, present illness, past medical history, and family history, and provide necessary, detailed, and accurate physical examinations, auxiliary tests, diagnoses, and treatment plans.  \\

\textbf{Consultation Steps:}\\
\textbf{Step 1:} Communicate with the patient to obtain their chief complaint, present illness, past medical history, and family history.  \\
\textbf{Step 2:} Once sufficient information is collected, recommend the necessary physical examination items. Physical examination refers to the process of assessing the patient's physical condition through observation, palpation, percussion, and auscultation to identify potential physical abnormalities.  \\
\textbf{Step 3:} After obtaining the results of the physical examination, recommend the necessary auxiliary tests based on the patient's chief complaint, present illness, past medical history, family history, and physical examination findings. Auxiliary tests include laboratory tests (e.g., blood tests, urine tests), imaging studies (e.g., X-rays, CTs, MRIs), or other specialized diagnostic methods to gather detailed information about the patient’s internal condition for diagnosis or disease evaluation.  \\
\textbf{Step 4:} After obtaining the auxiliary test results, provide a detailed and accurate diagnosis and treatment plan based on the patient's chief complaint, present illness, past medical history, family history, physical examination findings, and auxiliary test results.\\  

\textbf{Requirements:}\\
- Use conversational and simple language; keep questions short and clear.  \\
- Do not output questions, examination items, diagnoses, and treatment plans simultaneously.\\  
- Use layman’s terms for initial diagnoses, avoiding professional jargon.  \\
- Avoid repeating the patient’s information in your questions.  \\
- Collect details about the patient’s condition step by step.  \\
- Provide an initial diagnosis only after gathering sufficient information.  \\
- Answer any questions the patient may have.  \\
- Provide examination items first, followed by diagnoses and treatment plans.  \\
- Only provide diagnoses and treatment plans after receiving examination results.  \\
- Diagnose the patient’s condition in detail and specify the type and severity of the illness.  \\
- If a specific diagnosis cannot be made based on the available information, provide a list of suspected conditions.\\  
- Treatment plans should include further examinations, referrals, medication therapy, and non-drug therapies.  \\
- For medication therapy, specify the drug name, dosage, and purpose of the treatment. 
- Ask only one question at a time.  \\
- Areas not covered in the patient’s physical examination and auxiliary tests are considered normal.  \\

\textbf{Special Notes:}\\
- If the patient provides insufficient test results, do not continue asking for further test results. Instead, provide an initial diagnosis and treatment plan based on the current information and recommend additional tests to be completed.  \\

\textbf{Consultation Process:}\\
\textbf{1.} Start the consultation by saying: “Hello, what seems to be the problem?”  \\
\textbf{2.} When you believe you have enough information to recommend physical examination items, provide them with the phrase: “The following physical examinations are necessary:”  \\
\textbf{3.} When you believe you have enough information to recommend auxiliary tests, provide them with the phrase: “The following auxiliary tests are necessary:”  \\
\textbf{4.} When you believe you have enough information to make a detailed diagnosis, provide it with the phrase: “The following is the diagnosis and treatment plan:” \\ 
\textbf{5.} After providing the diagnosis and treatment plan, conclude with: “Consultation ended.”
\end{tcolorbox}

\clearpage

\supmethod{Model evaluation settings and development set design}{SM3}

\subsubsection*{Model inference settings and prompt design}
We employed the default parameter settings for each LLM, as LLMs demonstrate strong cross-task generalization capabilities and robustness to hyperparameters, making extensive hyperparameter tuning largely unnecessary. Tuning the hyperparameters of each model for each task was prohibitively expensive due to the substantial manual annotation required. Moreover, our objective was to assess the performance of LLMs in realistic work scenarios of GPs rather than to compare the models themselves. 
To ensure the validity of the benchmark results, no hyperparameters tuning was performed on the test set. Prompt design adhered to widely accepted practices, explicitly defining the model’s role, core capabilities, and task objectives. All experiments—including prompt design, model inference, and evaluation—were conducted in Chinese to reflect the authentic clinical linguistic environment. 
To evaluate the effectiveness and stability of the prompts, both automated and manual assessments were carried out on a small, independent development set, which included both objective and subjective tasks and excluded all test set instances. The finalized prompts were applied consistently across all models under evaluation. Advanced optimization techniques such as few-shot demonstrations, self-consistency, or iterative refinement were deliberately excluded to ensure that the evaluation results reflect the models’ inherent clinical reasoning capabilities.

\subsubsection*{Development set and test set}
We constructed a dedicated development set to support prompt validation and future experiments involving few-shot learning or hyperparameter tuning. The complete GPBench dataset comprises 88 standardized clinical cases encompassing 8 top common chronic diseases and 10 most common symptoms, along with 4,011 multiple-choice questions accompanied by verified reference answers. To ensure rigorous evaluation, the data were split as follows:

\begin{itemize}
    \item Development set: 
    \begin{itemize}
        \item MCQ Dev Set: 350 multiple-choice questions sampled using a stratified approach according to evaluation dimensions to ensure balanced coverage.
        \item Clinical Case Dev Set: 18 cases randomly selected from the total 88, balanced by disease category and common symptoms, used exclusively for prompt validation and hyperparameter experiments.
        \item AI Patient Dev Set: ``AI patient'' agents generated from the 18 clinical cases in the Clinical Case Dev Set following the designed prompt instructions.
    \end{itemize}

    \item Test set:
    \begin{itemize}
        \item MCQ Test Set: The remaining 3,661 multiple-choice questions.
        \item Clinical Case Test Set: The remaining 70 clinical cases, used exclusively for final evaluation.
        \item AI Patient Test Set: ``AI patient'' agents generated from the 70 clinical cases in the Clinical Case Test Set using the same prompt instructions.
    \end{itemize}
\end{itemize}
This design ensures a strict separation between development and test sets, enabling prompt validation and controlled tuning without compromising benchmark integrity.

\supmethod{Supplementary materials for human evaluations}{SM4}

To provide a reference for interpreting LLM performance in GPBench, we conducted a human evaluation involving general practitioners (GPs). All testing procedures were fully aligned with those applied to the LLMs, ensuring direct comparability. A total of six GPs participated in the evaluation, divided into three groups of two participants each to represent varying levels of clinical experience: two with more than 10 years, two with 5–10 years, and two with less than 5 years of general practice experience, as detailed in Supplementary Table~\ref{tab:human_group}. All assessments were conducted independently in a quiet environment, without the aid of any external tools.

\begin{table}[h]
    \centering
    {\fontsize{8pt}{8pt}\selectfont
    \caption{Basic information of the GPs who participated in the evaluation.}
    \label{tab:human_group}
\begin{tabular*}{0.9\textwidth}{@{\extracolsep{\fill}}lll@{}}
\toprule
Group & The Number of Participants & Years of Working Experience in General Practice
 \\ \midrule
A & 2 &$ >10$ years \\ 
B & 2 & $5 \sim 10$ years \\
C & 2 & $ <5$ years \\ 
\botrule
\end{tabular*}}
\end{table}

The test materials consisted of three components, detailed as follows:
\begin{itemize}
\item MCQ Test Set: A total of 200 multiple-choice questions were selected from the complete MCQ test set using a stratified random sampling method under the guidance of clinical physicians, ensuring balanced coverage across all 14 competency dimensions. Participants completed the questions via a digital questionnaire, and accuracy was automatically calculated.
\item Clinical Case Test Set: All 70 real-world outpatient cases were presented in a structured questionnaire format, requiring participants to provide free-text responses addressing both diagnosis and treatment plans.
\item AI Patient Test Set: The same 70 cases were deployed as interactive simulated consultations on the Dify platform. Participants interacted with AI-generated ``patients'' using standardized prompts and scripted dialogues identical to those applied to the LLMs.
\end{itemize}

Before testing, participants received standardized instructions consistent with the prompts used for the LLMs. The responses of GPs to the Clinical Case Test Set and the AI Patient Test Set were manually evaluated by experienced general practitioners using the same scoring criteria and procedure.
The detailed results of the human evaluations are presented in Supplementary Table 9.

\supmethod{ Supplementary materials for the Clinical Case Test Set and the AI Patient Test Set}{SM5}

For the Diagnostic Plan, input the patient's basic information, chief complaint, medical history, along with the results of physical and auxiliary examinations into the LLM. Based on the provided information, the LLM will assess whether a referral is necessary, determine if the case constitutes an emergency or critical condition, identify any existing complications, and produce a diagnostic result.

For the Treatment Plan, input the patient's basic information, chief complaint, medical history, physical examination findings, auxiliary examination results, and the confirmed diagnostic result. The LLM will generate an optimal treatment plan based on this information and will additionally recommend further examinations, health education, and measures related to controlling treatment costs.

For patient agents in the AI Patient test set, the input to the LLM consists of the patient's basic information, chief complaint, and medical history. The doctor agent, by contrast, receives no prior patient information and must acquire all relevant details through interaction with the patient agent. It should be noted that discrepancies may exist between the examination orders (both physical and auxiliary) issued by the doctor agent and the actual results. During the conversation, the true outcomes of the physical and auxiliary examinations are provided to the doctor agent. Based on the information gathered from the interaction and the examination results, the doctor agent formulates a diagnostic conclusion and proposes a treatment plan.

In the Clinical Case Test Set and AI Patient Test Set, the I3-1 indicator—formulation of optimal treatment plan—consists of two components: the optimal treatment plan and the clinical appropriateness evaluation of diagnostic examinations. Experts conducted fine-grained annotations for both components within the medical records. During evaluation, each component was independently scored by the three-member GP panel on a 100-point scale based on the expert-annotated standards. The final I3-1 score was then computed as the weighted average of the two component scores, with equal weighting applied.


\begin{thebibliography}{10}
\expandafter\ifx\csname url\endcsname\relax
  \def\url#1{\burl{#1}}\fi
\expandafter\ifx\csname urlprefix\endcsname\relax\def\urlprefix{URL }\fi
\providecommand{\bibinfo}[2]{#2}
\providecommand{\eprint}[2][]{\url{#2}}
\providecommand{\doi}[1]{\url{https://doi.org/#1}}
\bibcommenthead

\bibitem{DH1}
\bibinfo{author}{Menezes, M. C.~S.} \emph{et~al.}
\newblock \bibinfo{title}{The potential of generative pre-trained transformer 4 ({GPT-4}) to analyse medical notes in three different languages: a retrospective model-evaluation study}.
\newblock \emph{\bibinfo{journal}{The Lancet Digital Health}} \textbf{\bibinfo{volume}{7}}, \bibinfo{pages}{e35--e43} (\bibinfo{year}{2025}).

\bibitem{DH2}
\bibinfo{author}{Bellini, V.} \& \bibinfo{author}{Bignami, E.~G.}
\newblock \bibinfo{title}{Generative pre-trained transformer 4 ({GPT-4}) in clinical settings}.
\newblock \emph{\bibinfo{journal}{The Lancet Digital Health}} \textbf{\bibinfo{volume}{7}}, \bibinfo{pages}{e6--e7} (\bibinfo{year}{2025}).

\bibitem{med-palm}
\bibinfo{author}{Singhal, K.} \emph{et~al.}
\newblock \bibinfo{title}{Large language models encode clinical knowledge}.
\newblock \emph{\bibinfo{journal}{Nature}} \textbf{\bibinfo{volume}{620}}, \bibinfo{pages}{172--180} (\bibinfo{year}{2023}).

\bibitem{med-palm2}
\bibinfo{author}{Singhal, K.} \emph{et~al.}
\newblock \bibinfo{title}{Toward expert-level medical question answering with large language models}.
\newblock \emph{\bibinfo{journal}{Nature Medicine}} \bibinfo{pages}{1--8} (\bibinfo{year}{2025}).

\bibitem{strong2023chatbot}
\bibinfo{author}{Strong, E.} \emph{et~al.}
\newblock \bibinfo{title}{Chatbot vs medical student performance on free-response clinical reasoning examinations}.
\newblock \emph{\bibinfo{journal}{JAMA internal medicine}} \textbf{\bibinfo{volume}{183}}, \bibinfo{pages}{1028--1030} (\bibinfo{year}{2023}).

\bibitem{gilson2023does}
\bibinfo{author}{Gilson, A.} \emph{et~al.}
\newblock \bibinfo{title}{How does chatgpt perform on the united states medical licensing examination ({USMLE})? the implications of large language models for medical education and knowledge assessment}.
\newblock \emph{\bibinfo{journal}{JMIR medical education}} \textbf{\bibinfo{volume}{9}}, \bibinfo{pages}{e45312} (\bibinfo{year}{2023}).

\bibitem{medqa}
\bibinfo{author}{Jin, D.} \emph{et~al.}
\newblock \bibinfo{title}{What disease does this patient have? a large-scale open domain question answering dataset from medical exams}.
\newblock \emph{\bibinfo{journal}{Applied Sciences}} \textbf{\bibinfo{volume}{11}}, \bibinfo{pages}{6421} (\bibinfo{year}{2021}).

\bibitem{googleAMIE2025}
\bibinfo{author}{McDuff, D.} \emph{et~al.}
\newblock \bibinfo{title}{Towards accurate differential diagnosis with large language models}.
\newblock \emph{\bibinfo{journal}{Nature}} \bibinfo{pages}{1--7} (\bibinfo{year}{2025}).

\bibitem{xiaolan2025evaluating}
\bibinfo{author}{Chen, X.} \emph{et~al.}
\newblock \bibinfo{title}{Evaluating large language models and agents in healthcare: key challenges in clinical applications}.
\newblock \emph{\bibinfo{journal}{Intelligent Medicine}}  (\bibinfo{year}{2025}).

\bibitem{sarraju2023appropriateness}
\bibinfo{author}{Sarraju, A.} \emph{et~al.}
\newblock \bibinfo{title}{Appropriateness of cardiovascular disease prevention recommendations obtained from a popular online chat-based artificial intelligence model}.
\newblock \emph{\bibinfo{journal}{JAMA}} \textbf{\bibinfo{volume}{329}}, \bibinfo{pages}{842--844} (\bibinfo{year}{2023}).

\bibitem{zhu2023can}
\bibinfo{author}{Zhu, L.}, \bibinfo{author}{Mou, W.} \& \bibinfo{author}{Chen, R.}
\newblock \bibinfo{title}{Can the chatgpt and other large language models with internet-connected database solve the questions and concerns of patient with prostate cancer and help democratize medical knowledge?}
\newblock \emph{\bibinfo{journal}{Journal of translational medicine}} \textbf{\bibinfo{volume}{21}}, \bibinfo{pages}{269} (\bibinfo{year}{2023}).

\bibitem{aime}
\bibinfo{author}{Tu, T.} \emph{et~al.}
\newblock \bibinfo{title}{Towards conversational diagnostic artificial intelligence}.
\newblock \emph{\bibinfo{journal}{Nature}} \bibinfo{pages}{1--9} (\bibinfo{year}{2025}).

\bibitem{johri2025evaluation}
\bibinfo{author}{Johri, S.} \emph{et~al.}
\newblock \bibinfo{title}{An evaluation framework for clinical use of large language models in patient interaction tasks}.
\newblock \emph{\bibinfo{journal}{Nature medicine}} \textbf{\bibinfo{volume}{31}}, \bibinfo{pages}{77--86} (\bibinfo{year}{2025}).

\bibitem{augenstein2024factuality}
\bibinfo{author}{Augenstein, I.} \emph{et~al.}
\newblock \bibinfo{title}{Factuality challenges in the era of large language models and opportunities for fact-checking}.
\newblock \emph{\bibinfo{journal}{Nature Machine Intelligence}} \bibinfo{pages}{852--863} (\bibinfo{year}{2024}).

\bibitem{ng2025rag}
\bibinfo{author}{Ng, K. K.~Y.}, \bibinfo{author}{Matsuba, I.} \& \bibinfo{author}{Zhang, P.~C.}
\newblock \bibinfo{title}{{RAG} in health care: a novel framework for improving communication and decision-making by addressing {LLM} limitations}.
\newblock \emph{\bibinfo{journal}{Nejm Ai}} \textbf{\bibinfo{volume}{2}}, \bibinfo{pages}{AIra2400380} (\bibinfo{year}{2025}).

\bibitem{tran2025multi}
\bibinfo{author}{Tran, K.-T.} \emph{et~al.}
\newblock \bibinfo{title}{Multi-agent collaboration mechanisms: A survey of {LLMs}}.
\newblock \emph{\bibinfo{journal}{arXiv preprint arXiv:2501.06322}}  (\bibinfo{year}{2025}).

\bibitem{kim2024mdagents}
\bibinfo{author}{Kim, Y.} \emph{et~al.}
\newblock \bibinfo{title}{Mdagents: An adaptive collaboration of llms for medical decision-making}.
\newblock \emph{\bibinfo{journal}{Advances in Neural Information Processing Systems}} \textbf{\bibinfo{volume}{37}}, \bibinfo{pages}{79410--79452} (\bibinfo{year}{2024}).

\bibitem{multi_agent_npj}
\bibinfo{author}{Chen, X.} \emph{et~al.}
\newblock \bibinfo{title}{Enhancing diagnostic capability with multi-agents conversational large language models}.
\newblock \emph{\bibinfo{journal}{NPJ digital medicine}} \textbf{\bibinfo{volume}{8}}, \bibinfo{pages}{159} (\bibinfo{year}{2025}).

\bibitem{kim2025medical}
\bibinfo{author}{Kim, Y.} \emph{et~al.}
\newblock \bibinfo{title}{Medical hallucinations in foundation models and their impact on healthcare}.
\newblock \emph{\bibinfo{journal}{arXiv preprint arXiv:2503.05777}}  (\bibinfo{year}{2025}).

\bibitem{holt2024data}
\bibinfo{author}{Holt, S.}, \bibinfo{author}{Qian, Z.}, \bibinfo{author}{Liu, T.}, \bibinfo{author}{Weatherall, J.} \& \bibinfo{author}{van~der Schaar, M.}
\newblock \bibinfo{title}{Data-driven discovery of dynamical systems in pharmacology using large language models}.
\newblock \emph{\bibinfo{journal}{Advances in Neural Information Processing Systems}} \textbf{\bibinfo{volume}{37}}, \bibinfo{pages}{96325--96366} (\bibinfo{year}{2024}).

\bibitem{wonca2012definition}
\bibinfo{author}{{WONCA Europe}}.
\newblock \bibinfo{title}{The european definition of general practice/family medicine-2023 edition}.
\newblock \emph{\bibinfo{journal}{Barcelona: WONCA Europe}}  (\bibinfo{year}{2023}).
\newblock \urlprefix\url{https://www.woncaeurope.org/page/definition-of-general-practice-family-medicine}.

\bibitem{scherger2007preparing}
\bibinfo{author}{Scherger, J.~E.}
\newblock \bibinfo{title}{Preparing the personal physician for practice ({P4}): essential skills for new family physicians and how residency programs may provide them}.
\newblock \emph{\bibinfo{journal}{The Journal of the American Board of Family Medicine}} \textbf{\bibinfo{volume}{20}}, \bibinfo{pages}{348--355} (\bibinfo{year}{2007}).

\bibitem{mcclelland1973testing}
\bibinfo{author}{McClelland, D.~C.}
\newblock \bibinfo{title}{Testing for competence rather than for" intelligence."}.
\newblock \emph{\bibinfo{journal}{American psychologist}} \textbf{\bibinfo{volume}{28}}, \bibinfo{pages}{1} (\bibinfo{year}{1973}).

\bibitem{boyatzis1991competent}
\bibinfo{author}{Boyatzis, R.~E.}
\newblock \emph{\bibinfo{title}{The competent manager: A model for effective performance}}  (\bibinfo{publisher}{John Wiley \& Sons}, \bibinfo{year}{1991}).

\bibitem{wang2023cmb}
\bibinfo{author}{Wang, X.} \emph{et~al.}
\newblock \bibinfo{title}{Cmb: A comprehensive medical benchmark in chinese}.
\newblock \emph{\bibinfo{journal}{arXiv preprint arXiv:2308.08833}}  (\bibinfo{year}{2023}).

\bibitem{MedBench}
\bibinfo{author}{Liu, M.} \emph{et~al.}
\newblock \bibinfo{title}{Medbench: A comprehensive, standardized, and reliable benchmarking system for evaluating chinese medical large language models}.
\newblock \emph{\bibinfo{journal}{Big Data Mining and Analytics}}  (\bibinfo{year}{2024}).
\newblock \urlprefix\url{https://www.sciopen.com/article/10.26599/BDMA.2024.9020044}.

\bibitem{from_jp_1}
\bibinfo{author}{{Collaborators, GBD and others}}.
\newblock \bibinfo{title}{Global, regional, and national age--sex specific all-cause and cause-specific mortality for 240 causes of death, 1990--2013: a systematic analysis for the global burden of disease study 2013}.
\newblock \emph{\bibinfo{journal}{The Lancet}} \textbf{\bibinfo{volume}{385}}, \bibinfo{pages}{117--171} (\bibinfo{year}{2015}).

\bibitem{from_jp_2}
\bibinfo{author}{Zhou, M.} \emph{et~al.}
\newblock \bibinfo{title}{Cause-specific mortality for 240 causes in china during 1990--2013: a systematic subnational analysis for the global burden of disease study 2013}.
\newblock \emph{\bibinfo{journal}{The Lancet}} \textbf{\bibinfo{volume}{387}}, \bibinfo{pages}{251--272} (\bibinfo{year}{2016}).

\bibitem{from_jp_3}
\bibinfo{author}{Peng, W.} \emph{et~al.}
\newblock \bibinfo{title}{Trends in major non-communicable diseases and related risk factors in china 2002--2019: an analysis of nationally representative survey data}.
\newblock \emph{\bibinfo{journal}{The Lancet Regional Health--Western Pacific}} \textbf{\bibinfo{volume}{43}} (\bibinfo{year}{2024}).

\bibitem{hurst2024gpt4o}
\bibinfo{author}{Hurst, A.} \emph{et~al.}
\newblock \bibinfo{title}{{GPT}-4o system card}.
\newblock \emph{\bibinfo{journal}{arXiv preprint arXiv:2410.21276}}  (\bibinfo{year}{2024}).

\bibitem{openaio1}
\bibinfo{author}{Jaech, A.} \emph{et~al.}
\newblock \bibinfo{title}{{OpenAI} o1 system card}.
\newblock \emph{\bibinfo{journal}{arXiv preprint arXiv:2412.16720}}  (\bibinfo{year}{2024}).

\bibitem{gemini1.5}
\bibinfo{author}{{Gemini Team}}.
\newblock \bibinfo{title}{Gemini 1.5: Unlocking multimodal understanding across millions of tokens of context}.
\newblock \emph{\bibinfo{journal}{arXiv preprint arXiv:2403.05530}}  (\bibinfo{year}{2024}).

\bibitem{qwen2.5}
\bibinfo{author}{Yang, A.} \emph{et~al.}
\newblock \bibinfo{title}{Qwen2.5 technical report}.
\newblock \emph{\bibinfo{journal}{arXiv preprint arXiv:2412.15115}}  (\bibinfo{year}{2024}).

\bibitem{deepseekv3}
\bibinfo{author}{Liu, A.} \emph{et~al.}
\newblock \bibinfo{title}{Deep{S}eek-{V}3 technical report}.
\newblock \emph{\bibinfo{journal}{arXiv preprint arXiv:2412.19437}}  (\bibinfo{year}{2024}).

\bibitem{deepseekr1}
\bibinfo{author}{Guo, D.} \emph{et~al.}
\newblock \bibinfo{title}{Deep{S}eek-{R}1: Incentivizing reasoning capability in llms via reinforcement learning}.
\newblock \emph{\bibinfo{journal}{arXiv preprint arXiv:2501.12948}}  (\bibinfo{year}{2025}).

\bibitem{huatuogpto1}
\bibinfo{author}{Chen, J.} \emph{et~al.}
\newblock \bibinfo{title}{Huatuo{GPT}-o1, towards medical complex reasoning with llms}.
\newblock \emph{\bibinfo{journal}{arXiv preprint arXiv:2412.18925}}  (\bibinfo{year}{2024}).

\end{thebibliography}
\end{document}